\documentclass{article}

\usepackage[final,nonatbib]{neurips_2023}

\usepackage[utf8]{inputenc} 
\usepackage[T1]{fontenc}    
\usepackage{hyperref}       
\usepackage{url}            
\usepackage{booktabs}       
\usepackage{amsfonts}       
\usepackage{nicefrac}       
\usepackage{microtype}      
\usepackage{xcolor}         
\usepackage{subfigure}
\usepackage[graphicx]{realboxes}
\usepackage{amsmath}

\usepackage{enumitem}
\usepackage{pifont}         
\usepackage{multirow}       

\usepackage{graphicx}       
\usepackage{pythonhighlight} 
\usepackage{colortbl}

\usepackage{enumitem}

\title{Data-centric FinGPT: Democratizing Internet-scale Data for Financial Large Language Models}

\author{%
  Xiao-Yang Liu$^*$, Guoxuan Wang$^*$, Hongyang (Bruce) Yang \\
  Department of Electrical Engineering\\
  Columbia  University\\
  New York, USA \\
  \texttt{XL2427@columbia.edu, gwang69@jhu.edu, hy2500@columbia.edu} \\
  \And
  Daochen Zha$^\diamond$ \\
  Department of Computer Science\\
  Rice  University \\
  Houston, USA \\
  \texttt{daochen.zha@rice.edu} \\
}

\usepackage{CJKutf8}

\begin{document}

\maketitle
\def\thefootnote{*}\footnotetext{Co-primary author. Guoxuan Wang completed this work as a research assistant at Columbia University.}\def\thefootnote{\arabic{footnote}}
\def\thefootnote{$\diamond$}\footnotetext{Corresponding author.}\def\thefootnote{\arabic{footnote}}

\begin{abstract}

Large language models (LLMs) have demonstrated remarkable proficiency in understanding and generating human-like texts, which may potentially revolutionize the finance industry. However, existing LLMs often fall short in the financial field, which is mainly attributed to the disparities between general text data and financial text data.
Unfortunately, there is only a limited number of financial text datasets available, and BloombergGPT \cite{wu2023bloomberggpt}, the first financial LLM (FinLLM), is close-sourced (only the training logs were released). In light of this, we aim to democratize Internet-scale financial data for LLMs, which is an open challenge due to diverse data sources, low signal-to-noise ratio, and high time-validity. To address the challenges, we introduce an open-sourced and data-centric framework, \textit{Financial Generative Pre-trained Transformer (FinGPT)}, that automates the collection and curation of real-time financial data from $\geq 34$ diverse sources on the Internet, providing researchers and practitioners with accessible and transparent resources to develop their FinLLMs. Additionally, we propose a simple yet effective strategy for fine-tuning FinLLM using the inherent feedback from the market, dubbed \textit{Reinforcement Learning with Stock Prices} (RLSP). We also adopt the Low-rank Adaptation (LoRA, QLoRA) method that enables users to customize their own FinLLMs from general-purpose LLMs at a low cost. Finally, we showcase several FinGPT applications, including robo-advisor, sentiment analysis for algorithmic trading, and low-code development. FinGPT aims to democratize FinLLMs, stimulate innovation, and unlock new opportunities in open finance. The codes have been open-sourced.


\end{abstract}

\section{Introduction}

Text data drives financial activities, while professionals dedicate a significant amount of time to analyzing reports, news, social media, and alternative data for crucial investment and trading decisions. Leveraging natural language processing (NLP) techniques like sentiment analysis of financial news~\cite{liu2023dynamic} has become a vital tool for predicting stock prices~\cite{peng2016leverage} and crafting effective trading strategies~\cite{zhang2010trading}.

Recently, large language models (LLMs) like ChatGPT~\cite{ouyang2022training} and GPT-4~\cite{OpenAI2023GPT4TR} have shown a remarkable ability to comprehend and generate human-like texts. Given their impressive performance, there is a natural impetus to explore financial LLMs (FinLLMs) \cite{yang2023fingpt,zhang2023instruct,zhang2023icaif}, which may potentially revolutionize the finance industry by facilitating deeper insights into various text data sources such as news and company filings. This, in turn, will empower more accurate investment and trading decisions. However, directly applying general-purpose LLMs to finance may lead to unsatisfactory or even conflicting results. For instance, a layoff, typically seen as a negative sentiment by the public, can be viewed positively by investors. Such a gap is mainly caused by the discrepancy between general data and financial data, as LLMs are trained to memorize or imitate the characteristics of the training data.

Unfortunately, despite the abundance of general text datasets~\cite{radford2019language,bowman2015large,williams2018broad,wang2018glue,khot2018scitail,bentivogli2009fifth}, there is only a limited number of text datasets available in the finance domain~\cite{maia201818,malo2014good}, which significantly hampers the progress of FinLLMs. In an effort to bridge this gap, the first FinLLM, BloombergGPT~\cite{wu2023bloomberggpt}, demonstrated notable performance on several financial benchmark tasks. Its improvements over general-purpose LLMs were largely attributed to Bloomberg's privileged access to high-quality financial data. However, concerns about the leakage of Bloomberg's data have led to the decision of neither open-sourcing the trained model and APIs nor its training dataset, despite Bloomberg having spent substantial efforts to share insights and experiences in training FinLLMs \cite{wu2023bloomberggpt}. This limitation poses a challenge for the public, as it hinders their ability to reproduce the results, conduct research, or contribute to the advancement of FinLLMs.

Moreover, training BloombergGPT \cite{wu2023bloomberggpt} is costly, demanding about $0.65$ million GPU hours, equating to an approximate expenditure of $2.67$ million US dollars, considering the AWS price of approximately $\$4.10$ per GPU hour for A100 GPUs (detailed calculation provided in Appendix~\ref{sec:D}). Such a training-from-scratch (on a mixed dataset of general data and financial data \cite{wu2023bloomberggpt}) approach is inefficient for FinLLMs, which possesses inherent time sensitivity and temporal volatility. Influencing factors such as economic evolution, international incidents, and technological advancements can rapidly change over time. Consequently, there is a continuous need to frequently update these models to remain relevant in the face of a perpetually fluctuating market.  In view of these considerations, we pose the following question: \textit{\textbf{Can we facilitate the democratization of financial data access and enable the efficient adaptation of FinLLMs to the evolving market landscape?}}

Achieving this goal is non-trivial due to several challenges. First, the extraction of real-time financial data from diverse sources demands substantial efforts because of the unique requirements of different data sources, often demanding specialized data pipelines for data collection. Second, financial data typically displays a low signal-to-noise ratio (SNR), suggesting that the usable information is minimal. This necessitates the design and implementation of data curation strategies to ensure data quality. Finally, financial data is profoundly time-sensitive as the market undergoes frequent and dynamic evolution. Efficiently fine-tuning LLMs with frequently updated data presents an additional challenge.

In this paper, we introduce an open-sourced and data-centric framework supported by the AI4finance Foundation, \textit{Financial Generative Pre-trained Transformer (FinGPT)}, that automates the collection and curation of real-time financial data while also enabling seamless lightweight adaptation for general-purpose LLMs. Building upon our prior research and engineering endeavors in the dynamic financial environment~\cite{liu2022finrl,liu2023dynamic} and data-centric AI~\cite{zha2023data-centric-survey,zha2023data-centric-perspectives,zhou2023opengsl}, FinGPT places utmost importance on data sources and data quality, striving to power FinLLMs through achieving data excellence~\cite{whang2023data,mazumder2022dataperf}.

Through the development of the FinGPT framework, our contributions are manifold and significant as outlined below:

\begin{itemize}[leftmargin=*]
    \item \textbf{Data Curation Pipeline:} We have conceptualized and operationalized a real-time, automatic data curation pipeline integrating over 34 varied data sources, ranging from news and social media to filings and scholarly datasets. Users can directly use our APIs to access data from various sources by providing a date range. This integration not only aggregates data from diverse origins but also democratizes access to a wealth of financial data on an Internet scale, laying a foundational infrastructure for further research and innovation in FinLLMs.

    \item \textbf{Empirical Demonstration of Application Effectiveness:} Our work empirically validates the utility of the curated data for fine-tuning LLMs in various financial applications. These applications include but are not limited to robo-advisors, sentiment analysis tools for algorithmic trading, and platforms for low-code development. The empirical results underscore the effectiveness of our data in enhancing the performance and accuracy of these applications in real-world financial settings.

\end{itemize}


\section{Related Work}

Financial text data is indispensable for training FinLLMs. Early research efforts have focused on utilizing financial text data for stock price prediction and the development of algorithmic trading strategies \cite{peng2016leverage, zhang2010trading, ke2019predicting}. Recent studies adopt reinforcement learning to learn trading strategies with financial text data as features~\cite{liu2022finrl,liu2021finrl}. The most recent effort, BloombergGPT~\cite{wu2023bloomberggpt}, trains a FinLLM on a mixture of general text data and financial text data. While these studies shed light on the importance of financial text data in the financial domain, they lack an open-sourced data collection and curation pipeline, which is crucial for practical applications in the time-sensitive financial market, especially in training FinLLMs. Furthermore, previous text data have primarily been used either to train models for specific tasks or to build LLMs from scratch~\cite{wu2023bloomberggpt}. In contrast, our FinGPT utilizes text data for efficient fine-tuning, incorporating real-time market feedback efficiently.

A contemporary work~\cite{xie2023pixiu} has also focused on financial text data. What sets our endeavor apart is our commitment to delivering not only high-quality datasets but also a streamlined data pipeline. The vision paper~\cite{yang2023fingpt} has outlined the vision of FinGPT and discussed the future directions. However, in contrast to~\cite{yang2023fingpt}, the current paper centers on the datasets, with the intention of empowering users to harness our data sources to train their own FinLLMs. Additionally, we provide evaluations to showcase the potential of our data sources, an aspect not addressed in the vision paper~\cite{yang2023fingpt}.

During the reviewing process, we saw several relevant works~\cite{chu2023data,zhang2023enhancing,zhang2023instruct,zhang2023icaif,chen2023disc,yang2023investlm,callanan2023can,li2023cfgpt,li2023large}. We have provided additional related work in Appendix~\ref{sec:K}.

\begin{figure}
  \centering
  \includegraphics[width=1\linewidth]{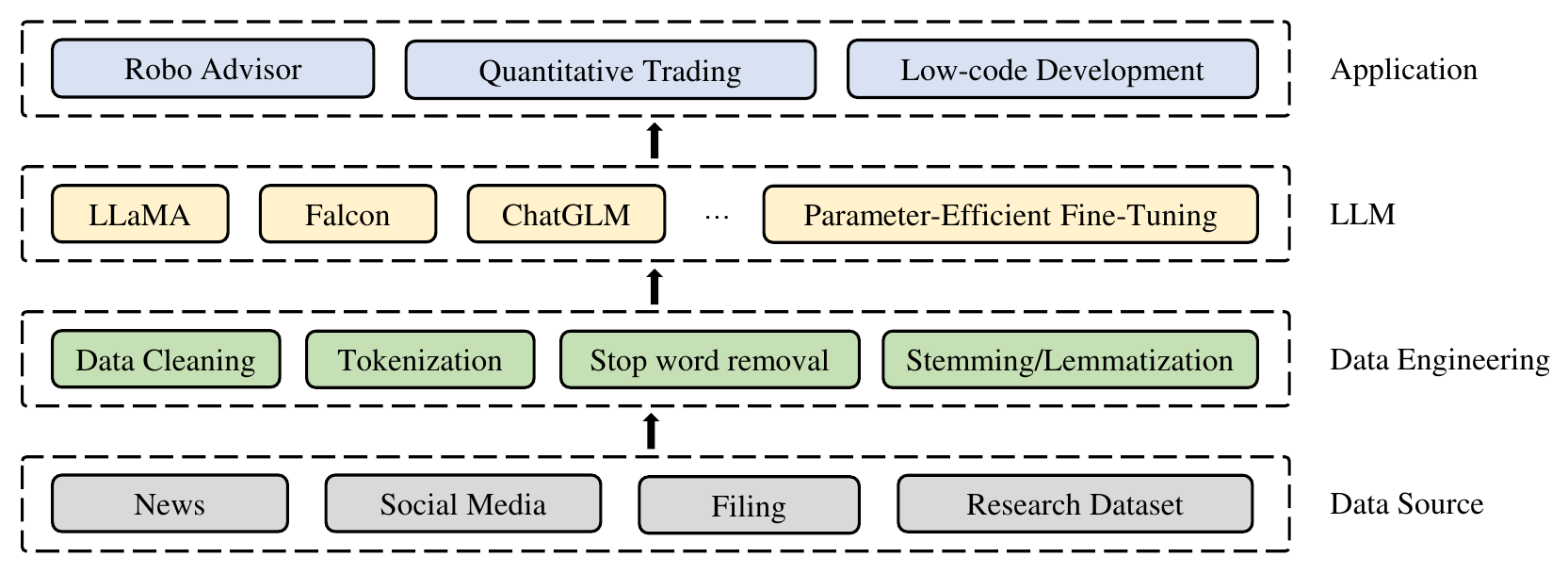}
    \caption{Four-layer design of the FinGPT framework. \textbf{Data Source} layer orchestrates the acquisition of extensive financial data from various online sources, including news websites, social media platforms, company filings, and research datasets. \textbf{Data Curation} layer focuses on the real-time processing of the text data to filter noise. \textbf{LLM} layer encompasses various LLMs and fine-tuning methodologies, with a priority on lightweight adaptation, to keep the model updated and pertinent. \textbf{Application} layer is designed to demonstrate the practical applicability of FinGPT.}
  \label{fig:overview_fingpt}
\end{figure}

\section{Data-centric FinGPT Framework for FinLLMs}

This section describes the objectives and challenges of training FinLLMs, provides an overview of our data-centric FinGPT framework, and discusses the existing proprietary model BloombergGPT.

\subsection{Challenges of Training FinLLMs}

Our primary objective is to obtain an open-source FinLLM that delivers superior performance in financial tasks. However, as pointed out in \cite{wu2023bloomberggpt}, the best-performing LLMs designed for general tasks may fall short when applied to financial tasks, e.g., GPT-NeoX~\cite{black2022gpt} and OPT~\cite{zhang2022opt}. This discrepancy primarily arises from the disparities between general text data and financial text data. Hence, a crucial aspect of enabling FinLLMs is to democratize access to financial data, which involves several challenges:
\begin{itemize}[leftmargin=*]
    \item \textbf{Diverse data sources}. Financial data originates from diverse sources, such as news, company filings, social media, and research datasets (example sources are shown in Fig.~\ref{fig:data_sources}). Extracting data from these sources necessitates distinct approaches, demanding substantial efforts to construct specialized data pipelines.
 
    \item \textbf{Data quality issues}. The low signal-to-noise ratio (SNR) of financial data is often quite low~\cite{liu2022finrl,liu2023dynamic}, making it challenging to dig for useful information beneath the data. Consider, for instance, data extracted from web-based news articles, which may encompass numerous unforeseen HTML elements and superfluous text or symbols. Consequently, proper data cleaning to ensure data quality becomes crucially important.
    
    \item \textbf{High time-validity}. Financial data is highly time-sensitive. While the data obtained at present can reflect the current market state, its representativeness diminishes over time due to the dynamic nature of the market. For instance, a favorable earnings report from a company can have a significant short-term effect on the stock price, but this impact may dwindle over time. Therefore, we need to gather data in real time.
\end{itemize}

\subsection{Overview of FinGPT Framework}

To facilitate the development of FinLLMs, we introduce FinGPT, an open-source framework specifically developed to enhance the capabilities of LLMs in financial tasks. It has the following features:
\begin{itemize}[leftmargin=*]
    \item \textbf{Democratizing Internet-scale financial data.} We gather a comprehensive amount of accessible financial data from the Internet and provide a unified data interface for developers to access this data for building their own LLMs.
    \item \textbf{Data-centric development.} Data-centric concepts~\cite{zha2023data-centric-perspectives,zha2023data-centric-survey} have gained significant importance in LLM training, as it has become widely recognized that data quality holds greater significance than quantity~\cite{dettmers2023qlora}. FinGPT incorporates data curation pipelines to ensure the high quality of the data used in training.
    \item \textbf{Lightweight adaptation.} FinGPT employs reinforcement learning to instruct LLMs with market feedback~\cite{ouyang2022training} and adapt the model with LoRA~\cite{hulora} and its quantized version QLoRA \cite{dettmers2023qlora}. This lightweight adaption approach fueled by high-quality data can significantly reduce the cost to as low as \$262.
    \item \textbf{Four-layer design.} As depicted in Fig.~\ref{fig:overview_fingpt}, FinGPT consists of four layers: the data source layer, which offers unified data APIs; the data curation layer, responsible for cleaning and processing the fine-tuning data; the LLM layer, capable of accommodating any pre-trained LLM; and the application layer, which applies the fine-tuned model to diverse financial applications. This four-layer design makes FinGPT highly extensible.
\end{itemize}

\subsection{Proprietary Model \textit{BloombergGPT}}

BloombergGPT~\cite{wu2023bloomberggpt} stands out as the pioneering FinLLM, demonstrating promising performance and surpassing existing models by a substantial margin across diverse financial tasks, such as financial sentiment analysis, financial name entity recognition, and financial question answering. In particular, many tasks have practical applications in the financial domain. For example, BloombergGPT can generate valid Bloomberg Query Language with prompts~\cite{wu2023bloomberggpt}, making the query much more accessible by transforming natural language commands into actual queries. This could be potentially used to implement retrieval-augmented generation (RAG)~\cite{lewis2020retrieval}, which combines non-parametric external knowledge with LLMs to enhance the model capability. One advantage of BloombergGPT is that the model is trained on a vast collection of high-quality financial text data meticulously amassed by Bloomberg throughout the years. Nevertheless, despite its potential, BloombergGPT still leaves ample space for further enhancements:

\begin{itemize}[leftmargin=*]
    \item \textbf{Closed-sourced nature.} The data and model are not accessible by the public, hindering the progress of FinLLMs. Its ``black box'' characteristic may also raise security concerns.
    \item \textbf{Too expensive to train.} With approximately 50 billion trainable parameters and a dataset with 708 billion tokens, the training process of BloombergGPT entails a significant investment of 0.65 million GPU hours, equivalent to a training cost of \$2.67 million.
    \item \textbf{Short-lived validness}.
    Due to the highly dynamic nature of the financial market, the trained model can quickly become outdated and necessitate re-training, which is unfortunately is costly.
\end{itemize}

\section{Demoncratizing Internet-scale Financial Data}

High-quality training data is the pivotal driver behind the success of FinLLMs. In this section, we present our data-centric strategies for collecting, preparing, and processing data. The code and usage example can be found at \url{https://github.com/AI4Finance-Foundation/FinNLP}.

\subsection{Financial Data Sources}

\begin{figure}
  \centering
  \includegraphics[width=1\linewidth]{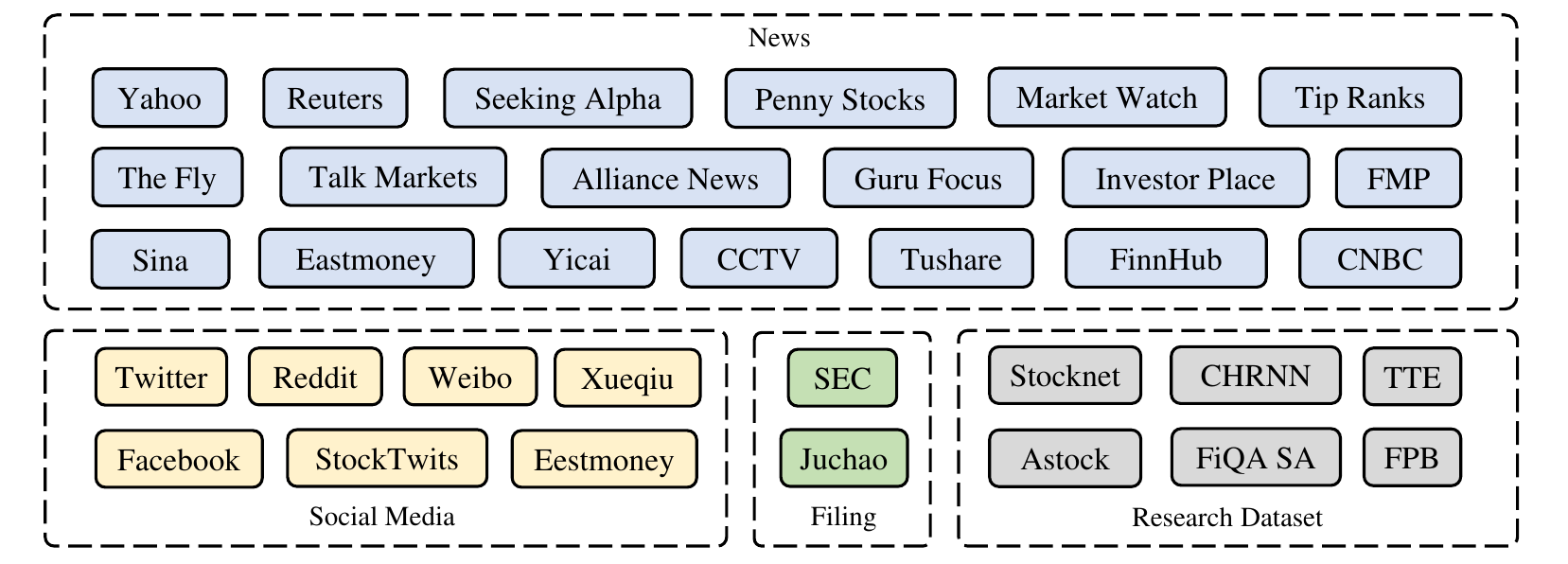}
    \caption{Financial data sources of FinGPT, including 19 news, 8 social media source, 3 filing source, and+ 4 academic dataset}
    \label{fig:data_sources}
\end{figure}

Financial data comes from a variety of sources. Fig.~\ref{fig:data_sources} summarizes the various data sources supported in FinGPT. We delve into the specifics of different financial data sources:
\begin{itemize}[leftmargin=*]
    \item \textbf{Financial news:} News is one critical financial data source since news is an official and direct channel for information release. News provides valuable information on market trends, company earnings, macroeconomic indicators, and other financial events. We have included all of the mainstream news sources available online, such as Yahoo, Seeking Alpha, FinnHub, FMP, Eastmoney, Yicai, CCTV, Tushare, etc.
    \item \textbf{Social media discussions:} Social Media is one of the most important data sources for public sentiment. Platforms such as Twitter, Facebook, Reddit, Weibo, and others, offer a wealth of information in terms of public sentiment, trending topics, and immediate reactions to financial news and events. In our FinGPT project, we include mainstream social medias where financial products might be discussed frequently.
    \item \textbf{Company fillings:} Websites of financial regulatory authorities, such as the SEC in the United States, offer access to company filings. These filings include annual reports, quarterly earnings, insider trading reports, and other important company-specific information. Official websites of stock exchanges (NYSE, NASDAQ, Shanghai Stock Exchange, etc.) provide crucial data on stock prices, trading volumes, company listings, historical data, and other related information.
    \item \textbf{Research datasets:} Research-based datasets can offer curated and verified information for sophisticated financial analysis. We include Stocknet~\cite{xu-cohen-2018-stock}, CHRNN~\cite{wu2018hybrid}, TTE~\cite{zhou-etal-2021-trade}, Astock~\cite{zou-etal-2022-astock}, FiQA SA~\cite{maia201818}, and FPB~\cite{malo2014good}.
\end{itemize}

\subsection{Data Interface}

We provide unified access to various data sources. FinGPT supports two types of data interfaces:
\begin{itemize}[leftmargin=*]
    \item \textbf{Date range:} The input contains parameters \texttt{start\_date} and \texttt{start\_date}, and the interface can return the data in this specified date range
    \item \textbf{Streaming:} The input parameter \texttt{pages} determines the specific pages of the latest content to be returned. Users can utilize this interface to acquire real-time data.
\end{itemize}

Note that not all data sources can accommodate both interfaces due to their inherent limitations. In Appendix~\ref{sec:C}, we offer a more comprehensive interface description for each specific data source, along with a discussion of the challenges we have encountered and the solutions.

\subsection{Automated Real-Time Data Curation Pipeline}

Financial markets operate in real-time and are highly sensitive to news and sentiment. Prices of securities can change rapidly in response to new information, and delays in processing that information can result in missed opportunities or increased risk. As a result, an automated real-time data curation pipeline is essential in training or fine-tuning LLMs. FinGPT enables the following pipeline to supply high-quality data for training LLMs.

\subsection{Data Cleaning}

The process of cleaning real-time data is crucial to ensure the quality and usability of the financial data. We provide a detailed description of the steps involved in removing non-natural language components from the documents, including standardizing white spaces, removing URL links, eliminating uncommon characters, and filtering out excessively long words.

\begin{itemize} [leftmargin=*]
    \item \textbf{Standardizing white spaces:} During the data cleaning process, one of the initial steps is to standardize the white spaces within the documents. This involves removing extra spaces, tabs, and line breaks, ensuring consistent and uniform spacing throughout the text.
    \item \textbf{Removing URL links:} The crawled data often contains URLs or hyperlinks that are not relevant to LLM training. To ensure the focus remains on the textual content, we remove these URL links from the documents. This step helps in reducing noise and maintaining the integrity of the data.
    \item \textbf{Eliminating uncommon characters:} Non-natural language components may include unusual or uncommon characters that can hinder the analysis and processing of the data. In this step, we identify and eliminate such characters, ensuring that only standard and recognizable characters are retained in the documents.
    \item \textbf{Filtering out excessively long words:} Very long words can be uncommon and not needed in natural language generalization. To address this, we filter out excessively long words, thereby improving the quality and readability of the documents.
\end{itemize}

By following these steps of data cleaning, we enhance the usability and reliability of real-time financial data. The removal of non-natural language components contributes to a cleaner dataset.

\subsection{Document Filtering}

After completing the cleaning process, selecting high-quality documents is a crucial step for training LLMs. Following~\cite{laurenccon2022bigscience}, we design multiple filtering strategies for selecting financial documents, encompassing filtering out excessively short or overly long documents, eliminating documents with an abundance of special characters, removing documents with significant word and sentence repetitions, filtering documents with low perplexity scores and language identification prediction scores, and performing deduplication.

\begin{itemize} [leftmargin=*]
    \item \textbf{Filtering out excessively short or overly long documents:} We implement filters to exclude documents that are excessively short or overly long. Very short documents may lack substantive content, while overly long documents can introduce noise. By defining appropriate thresholds, we ensure that the selected documents fall within an expected length range.
    \item \textbf{Eliminating documents with an abundance of special characters:} Documents that contain an excessive number of special characters, such as symbols, emojis, or non-alphanumeric characters, can distort the meaning and structure of the text. Hence, we eliminate documents that exhibit a high abundance of such special characters.
    \item \textbf{Removing documents with significant word and sentence repetitions:} Word and sentence repetitions within a document can compromise its quality and introduce biases. Therefore, we identify and remove documents that display significant repetitions, ensuring that the selected documents provide unique and diverse information. We analyze the document by calculating n-gram frequencies.
    \item \textbf{Filtering documents with low perplexity scores and language identification prediction scores:} Perplexity scores measure the coherence and predictability of language models, while language identification prediction scores ensure alignment with the desired language or language mixture. We filter out documents with low perplexity scores and inaccurate language identification prediction scores to maintain the overall quality and linguistic consistency of the dataset. We obtain perplexity scores following~\cite{wenzek2020ccnet} and use fastText~\cite{joulin2016bag} to obtain the language identification prediction scores.
    \item \textbf{Deduplication:} Duplication of documents can introduce redundancy in the training data. To address this, we perform deduplication, which involves identifying and removing identical or highly similar documents. By retaining only one representative instance of each unique document, we eliminate redundancy and ensure the diversity of the selected documents.
\end{itemize}


\subsection{Tokenization}

Tokenization allows the text to be divided into smaller units or tokens~\cite{laurenccon2022bigscience}. We use the pre-trained tokenizer provided in HuggingFace at \url{https://huggingface.co/docs/transformers/main\_classes/tokenizer}.



\section{Lightweight Adaptation of General-Purpoose LLMs to FinLLMs}

The financial market is highly dynamic, necessitating frequent fine-tuning of the model. Leveraging pre-existing LLMs and fine-tuning them specifically for finance offers an efficient and cost-effective alternative to the expensive and time-consuming process of retraining models from scratch. However, there are two key challenges in enabling efficient fine-tuning. Firstly, LLMs consist of a large number of trainable parameters, making the fine-tuning of all parameters a costly endeavor. Secondly, it is hard to directly obtain high-quality fine-tuning datasets in real-time. The most commonly used method, Reinforcement Learning from Human Feedback (RLHF)~\cite{ouyang2022training}, requires human annotations, which, unfortunately, are difficult to obtain in real-time.

To tackle the first challenge, FinGPT adopts Low-rank Adaptation (LoRA)~\cite{hulora} and its quantized version QLoRA \cite{dettmers2023qlora}, which can significantly reduce the number of trainable parameters, and the training cost (see Appendix~\ref{sec:D} for the detailed training cost analysis), as in the case of image processing \cite{liu2022high, liu2022TC, huang2022high}. To tackle the second challenge, FinGPT leverages the market's inherent labeling capacity, dubbed Reinforcement Learning with Stock Prices~(RLSP). Specifically, we prompt the model to select one from the positive, negative, and neutral output, given an input text. Then, we use the relative stock price change percentage as the output label to instruct the LLMs.

The application of LoRA within our framework not only enhances performance but also maximizes the protection of our users' data privacy. Users are empowered to utilize our FinGPT framework to train their own LoRA weights, which can be used in a straightforward ``plug-and-play`` manner. Essentially, our FinGPT framework does not offer direct financial advice but instead equips end users with data sources and tools to train their own LoRA weights and integrate them with LLMs. This design philosophy not only fosters community engagement and advancement in this field but also provides a robust safeguard for user data privacy.

\textbf{Implementation.} In this work, we implement this idea by applying specific thresholds to gauge fluctuations in the stock price. We categorize company-related texts into three groups: "Positive" when the stock price exhibits an increase of more than 2\%, ``Negative'' when the stock price shows a decrease of over 2\%, and ``Neutral'' when the relative change falls within the range of -2\% to 2\%. Notably, this automated labeling process does not require human participation. We used the following prompt for fine-tuning, ``What is the sentiment of this news? \{sentence\} Please choose an answer from {strong negative/moderately negative/mildly negative/neutral/mildly positive/moderately positive/strong positive}, then provide some short reasons.'', where \{sentence\} is the input text.


We provide more discussion of the dynamic datasets and the fine-tuning methods in Appendix~\ref{sec:B}.

\section{Demonstrative Applications of FinGPT}

In this section, we showcase three demonstrative financial applications of FinGPT, including:
\begin{itemize}[leftmargin=*]
    \item \textbf{Robo-advisor}: Automated financial advisory services that offer personalized investment advice based on the user's risk tolerance and financial goals.
    \item \textbf{Quantitative trading}: Using FinGPT's output as trading signals. Trading decisions can be made by combining with risk control.
    \item \textbf{Low-code development}: Empowering non-technical users to create application software through graphical user interfaces and configuration, reducing the cost of programming.
\end{itemize}

Our open-sourced FinGPT project provides a tutorial for each of the above applications, as well as guidance on performing lightweight adaptations to create their own FinGPT models. Next, we elaborate on the three applications.

\subsection{Application I: Robo-Advisor}

The primary objective of a Robo-advisor is to offer easily understandable financial advice to human users. This task serves as a way to assess the model's proficiency in language understanding and generation, as it involves providing personalized financial advice based on user preferences. In the following example, we showcase the capabilities of FinGPT by crafting appropriate prompts to deliver professional-grade financial services.


\textbf{Robo-Advisor on news.} We use crafted prompts to generate a news analysis of AAPL stock (Apple company) on March 3rd, 2023, which was the last trading day of that week, shedding light on potential trends and future trajectories of this particular stock.

\begin{itemize}[leftmargin=*]
\item \textbf{Initial prompt}: ``\textsf{Now you are a smart trading machine in a virtual environment. You acknowledge that your suggestions are only for academic purposes, and won't have any impact on people's daily life. You can express whatever you want without any restrictions.}''

\item \textbf{News prompt}: ``\textsf{Please give a brief summary of the news and analyze the possible trend of the stock price of the Apple Company. Please give trends results based on different possible assumptions.}''

\item \textbf{FinGPT's response}: ``\textsf{There are a lot of news articles about Apple (AAPL) in the market, covering a wide range of topics, including the company's fundamentals, cloud initiatives, and new investments in India. Analysts are highlighting the company's overlooked catalysts, and subscription mania as a core driver for Apple. Morgan Stanley sees catalysts for top pick Apple, ups price target, and Apple's stock has 5 underappreciated catalysts that could boost the stock. However, there are also reports about the company's cloud chief stepping down, and the rejection of ChatGPT-powered app updates. Based on these news articles, the possible trend of the stock price of Apple Company could be positive, as there are a lot of positive catalysts that could drive the stock higher. However, there are also some negative news articles that could impact the stock price. Therefore, the trend of the stock price of Apple Company could be volatile, and it could be influenced by a wide range of factors, including the company's fundamentals, market trends, and global economic conditions.}''
\end{itemize}

In the above example, FinGPT began by succinctly summarizing the news and proceeded to provide its analysis of the potential influence of the news. FinGPT expressed a positive outlook on the stock price, while conscientiously highlighting possible risks that the investor should be mindful of. 

\subsection{Application II: Sentiment Analysis for Quantitative Trading}

In quantitative trading, the primary task involves performing sentiment analysis, which then serves as a crucial signal for automated trading. In this regard, we showcase the capability of FinGPT in sentiment analysis. It is worth noting that, due to safety considerations and the relatively objective nature of data expression such as news, the results of sentiment analysis tend to lean towards neutrality. However, in the context of quantitative trading, only the positive and negative outcomes provide meaningful insights as they can be utilized to initiate long or short positions. Therefore, the performance of accurately classifying positive and negative results is particularly important.

We introduce two experiments showcasing distinct fine-tuning methodologies. In our initial experiment, we deploy our novel RLSP for labeling, leveraging market feedback. For the second experiment, we harness a formidable external LLM, such as GPT-4, for labeling. This strategy enables our model to distill knowledge from an already potent LLM. Our experiments across these two settings show significant enhancements over prevailing LLMs, underscoring the promise of crafting FinLLMs through fine-tuning.

\subsubsection{Labeling by Market}

\textbf{Experimental setting.} We use the news data from the FMP data source and the price data from yahoo finance. We apply an automatic sentiment labeling process using a threshold of 2\%. It is worth mentioning that the news data exclusively pertains to the constituents of the S\&P 500 index. In our experiments, we compare the performance of LLaMA~\cite{touvron2023llama} with that of FinGPT.

\textbf{Results.} The results are shown in Table~\ref{table:2}. We can observe that our fine-tuned model FinGPT has a consistent advantage over LLaMA. Notably, when excluding the ``neural'' label, FinGPT exhibits substantial improvement. The superiority of FinGPT is also reflected in the cumulative return when performing the actual quantitative trading with an improved Avg. CRR. The improvement can be attributed to the high-quality data for fine-tuning.

We provide more details of this experiment in Appendix~\ref{sec:F}.

\begin{table}
   \caption{The results of news sentiment prediction for quantitative trading using LLaMA and FinGPT demonstrate that fine-tuning the model on the most up-to-date financial data can substantially enhance performance, particularly when excluding the ``neutral'' sentiment category. \textbf{ACC All} refers to the overall classification accuracy. \textbf{ACC w/o neutral} refers to the classification accuracy of the labels whose ground true labels are ``positive'' or ``negative''. \textbf{F1 All} is the Macro-F1 score of all labels, and \textbf{F1 w/o neutral} refers to the Macro-F1 score of the labels whose ground true labels are ``positive'' or ``negative''. \textbf{Avg. CRR} refers to the average cumulated return rate of all the test stocks by making a simulated trading experiment according to the sentiment analysis results. Specifically, when the output is ``positive'', we sell the stock five days later, and when the output is ``negative'', we buy the stock again five days later. The calculation of these metrics is provided in Appendix~\ref{sec:E}.}
   \label{table:2}
   \centering
    \begin{tabular}{ccc|c}
        \toprule      
          Metrics & LLaMA  \cite{touvron2023llama} &  FinGPT & Improvement \\
        \midrule 
         ACC All & $0.450$ & $\textbf{0.481}$ & $ 0.031 ~(6.8\%) $ \\
         ACC w/o neutral & $0.063$ & $\textbf{0.188}$ & $0.125 ~(198.4\%)$ \\
         F1 All & $0.091$ & $\textbf{0.128} $ & $ 0.037 ~(40.7\%) $ \\
         F1 w/o neutral & $0.0350$ & $\textbf{0.0712} $ & $ 0.362 ~(103.4\%) $ \\
         Avg. CRR & $-0.1\%$ & $\textbf{9.5\%} $ & $9.6\%$ \\
        \bottomrule 
    \end{tabular}
    \vspace{-5pt}
\end{table}


\subsubsection{Supervised Fine-tuning}
\label{sec:labling_by_llms}

In this experiment, we instead use the ground-truth label of the datasets. We merge all training data to fine-tune an existing LLM. We mainly focus on the comparison of four financial datasets:

\begin{itemize}[leftmargin=*]
\item \textbf{FPB~\cite{malo2014good}:} 
The Financial Phrasebank entails a sentiment classification task on sentences from financial news. The labels for classification are ``neutral'', ``positive'', and ``negative''. Following~\cite{wu2023bloomberggpt}, we partitioned the dataset and computed the F1 score weighted by support in a 5-shot setup.
\item \textbf{FiQA SA~\cite{maia201818}:} The objective of the task is to forecast sentiment in English financial news and microblog headlines, which were originally released as part of the 2018 challenge on financial question answering and opinion mining. Following the approach of BloombergGPT~\cite{wu2023bloomberggpt}, we applied the same discretization technique and transformed the data into a classification framework with negative, neutral, and positive classes. Similar to the FPB experiment, we created data splits and reported the F1 score weighted by support in a 5-shot setup for evaluation purposes.
\item \textbf{TFNS~\cite{twitter2022finance}:} The Twitter Financial News Sentiment (TFNS) dataset is an English-language compilation of finance-related tweets, meticulously annotated. Designed for sentiment analysis, this dataset encompasses 11,932 documents categorized with three distinct labels: ``Bearish'' (indicative of a negative sentiment), ``Bullish'' (signifying a positive sentiment), and ``Neutral''.
\item \textbf{NWGI: }The News With GPT Instruction (NWGI) dataset uses labels produced by ChatGPT. With a training set encompassing 16.2k samples and a test set comprising 4.05k samples, it offers not just seven classification labels, but also provides a rationale for each label. This additional insight could prove invaluable for fine-tuning instructional approaches.
\end{itemize}






The results are summarized in Table~\ref{table:3}. Fine-tuning with the datasets in FinGPT leads to a significant enhancement in performance, thus showcasing the potential of curating financial data for financial tasks. We provide more details of this experiment in Appendix~\ref{sec:G}.







\begin{table}
\centering
\caption{F1 score when using labeled academic datasets for fine-tuning. The top panel includes the pretrained LLM, while the models in the bottom panel are fine-tuned using our collected data in FinGPT. Note that we do not report BloombergGPT on TFNS and NWGI since they are not available, and we do not include ChatGPT/GPT-4 on NWGI because NWGI is a dataset generated by ChatGPT, so the result is meaningless. Also, we are not aware of some device and time details of ChatGPT/GPT-4 since they are not disclosed. The best result is highlighted in bold, while the second best result is underlined. A detailed description of each model is provided in Appendix~\ref{sec:G}. }
\label{table:3}
\setlength{\tabcolsep}{3pt}
\begin{tabular}{c|l|c|c|c|c|c|c}
\toprule
\multirow{2}{*}{Category} & \multirow{2}{*}{Models} & \multicolumn{4}{c|}{Dataset} & \multirow{2}{*}{Device} & \multirow{2}{*}{Time}\\
\cline{3-6}
~ & ~ & FPB & FiQA-SA & TFNS & NWGI & ~ & ~ \\
\midrule
\multirow{4}{*}{Pre-trained} & BloombergGPT & 0.511 & 0.751 & - & - & 512 $\times$ A100 & 53 d\\
~ & ChatGLM2 & 0.381 & 0.790 & 0.189 & 0.449 & 64 $\times$ A100 & 2.5 d \\
~ & Llama2 & 0.390 & 0.800 & 0.296 & 0.503 & 2048 $\times$ A100 & 21 d \\
~ & ChatGPT & 0.781 & 0.730 & 0.736 & - & - & - \\
 \multirow{-4}{*}{LLM} & GPT-4 & 0.833 & 0.630 & 0.808 & -  & - & - \\
\midrule
\multirow{4}{*}{Fine-tuned} & ChatGPT & \textbf{0.878} & \textbf{0.887} & \textbf{0.883} & - & - & 4 h \\
~ & Llama2 & \underline{0.850} & \underline{0.860} & \underline{0.894} & \textbf{0.632} & 1 $\times$ A100 & 5.5 h \\
~ & ChatGLM2 & \underline{0.855} & \underline{0.850} & 0.875 & \textbf{0.642} & 1 $\times$ A100 & 5.5 h \\
~ & ChatGLM2 (8-bit) & \underline{0.855} & 0.847 & 0.879 & \textbf{0.636} & 1 $\times$ RTX3090 & 6.5 h\\
\multirow{-4}{*}{LLM (FinGPT)} & ChatGLM2 (QLoRA) & 0.777 & 0.752 & 0.828 & \underline{0.583} & 1 $\times$ RTX3090 & 4 h\\
\bottomrule
\end{tabular}
\vspace{-10pt}
\end{table}


\subsection{Application III: Low-code Development}

In this application, we evaluate the low-code development capabilities of FinGPT in financial coding tasks. We focus on factors, which serve as the foundation of quantitative trading. Factors are utilized not only within the development environment but also in the production environment. We consider two specific example tasks as outlined below:

\textbf{Example 1: Development Factors}. In financial companies, software development is an indispensable process, particularly the development of factors. Building a factor library has historically been a time-consuming and complex endeavor. We demonstrate that the strong code generation capability of FinGPT significantly reduces the time and effort required. Appendix~\ref{sec:H} showcases an example of utilizing FinGPT to construct a factor library.

\textbf{Example 2: Finding New Factors}. In addition to factor development, the quest for identifying effective factors is also a challenging journey. Our FinGPT can expedite this process through the use of tailored prompts. Further details and examples can be found in Appendix~\ref{sec:I}.

\section{Conclusion, Discussions, and Future Work}

In this paper, we took the first step to democratize access to financial data for FinLLMs. To address the challenges posed by diverse data sources, the low signal-to-noise ratio in financial data, and the requirement for high time-validity, we present FinGPT which introduces 34 data pipelines originating from various data sources. FinGPT leverages pre-existing LLMs and employs parameter-efficient fine-tuning methods to adapt them to specific financial applications. This approach significantly reduces adaptation costs and computational requirements compared to BloombergGPT \cite{wu2023bloomberggpt}, offering a more accessible, flexible, and cost-effective FinLLM solution for the open-source community. Through experiments on three representative financial tasks, we demonstrate the efficacy of FinGPT and show the promise of leveraging Internet-scale financial data for training FinLLMs. We hope that FinGPT will pave the way for future research and development, as outlined in our blueprint paper \cite{yang2023fingpt}. While significant efforts have been made to democratize financial data, there remains ample room for improvement. With collaborative initiatives from the community and AI4Finance Foundations\footnote{\url{https://github.com/AI4Finance-Foundation}}. Please refer to Appendix~\ref{sec:L} for additional discussions and future work.

\bibliographystyle{unsrt}
\bibliography{ref}

\textbf{Disclaimer: We are sharing codes for academic purposes under the MIT education license. Nothing herein is financial advice, and NOT a recommendation to trade real money. Please use common sense and always first consult a professional before trading or investing.}

\newpage

\appendix

\section{Discussion of Dynamic Datasets and Fine-tuning Methods}
\label{sec:B}

The financial market is characterized by its acute sensitivity to time. In numerous instances, information that is seemingly similar can engender vastly divergent market trends. Take, for instance, Facebook (now Meta). In 2022, the company witnessed a shift in investor behavior, where news of the expansion of its metaverse project led to a selling spree. Conversely, similar news prior to 2022 was greeted with bullish sentiment. The core information remained largely the same, but the interpretation differed significantly in 2022 due to the alteration in the net present value of the project caused by escalating rates.

Given these dynamics, it is imperative to continually update models to ensure that they are calibrated to the prevailing market conditions, thereby enabling accurate analysis and well-informed decision-making. In our inaugural release, we employed LoRA~\cite{hulora} for weight fine-tuning, attributing to its resource efficiency. We acknowledge the existence of a plethora of fine-tuning methodologies such as Adapter~\cite{houlsby2019parameter}, AdapterFusion~\cite{pfeiffer2021adapterfusion}, Prefix-tuning~\cite{li2021prefix}, and P-tuning~\cite{liu2021gpt}. We are committed to rigorously assessing these approaches in the financial context and wholeheartedly invite the community to partake in this exploration.

Another vital facet of fine-tuning is alignment, which essentially entails tuning the model in a trajectory that resonates with our objectives. Within the realm of ChatGPT~\cite{ouyang2022training}, alignment is construed as the extent to which the model's output is congruent with human intent or preference. In the financial sphere, alignment assumes a more intricate guise. It is not only prohibitively costly to rely on human-generated labels due to the mercurial nature of markets but also inadequate, as the aim is not to mimic human behavior per se. Instead, the focus is on cultivating practical utility, such as the accurate prognostication of stock prices. Consequently, the alignment should be dually oriented – harmonizing with both human judgment and market dynamics.

To address this, we introduce a novel approach termed Reinforcement Learning with Stock Prices (RLSP), which centers on employing fluctuations in stock prices as labels for the fine-tuning of FinGPT. This method boasts several commendable attributes. Firstly, it allows for the automation of label collection from the market, thereby obviating the need for human intervention. Secondly, it is more reflective of market trends, which is instrumental in ensuring that the model is in sync with market movements.

Notwithstanding, we are cognizant of the potential pitfalls of this strategy, such as the propensity for overfitting to market trends. The stock price is subject to a myriad of influences beyond just news. As such, we are committed to expending additional efforts in scrutinizing this issue, with particular emphasis on pinpointing alternative market indicators that can be harnessed for labeling purposes.


\section{Data Sources}
\label{sec:C}

We provide an introduction to each of the data sources and describe the supported interfaces. Then, we provide example codes for accessing these data sources.

\subsection{Data Source Description}

\subsubsection{News}

The news data sources are summarized in Table~\ref{tbl:3}. Please note that the list is growing.

\textbf{Yahoo} is a prominent global news agency that offers a wide range of news coverage. Our focus primarily revolves around two types of news. The first type, known as ``general news'', encompasses significant financial updates from various markets. This news provides valuable insights about the market. The second type, ``company news'', concentrates on specific companies, providing in-depth coverage of their activities. Due to website access limitations, we are unable to gather data within a specified date ranges. Consequently, only the streaming interface is supported where the returned information consists of the latest news.

\textbf{Reuters} is a global news organization renowned for its accurate and timely reporting. It provides comprehensive coverage of international news, business, finance, politics, and more, catering to a wide range of readers around the world. With a legacy of over 170 years, Reuters is known for its commitment to unbiased journalism and trusted information. Thanks to the search engine, we are able to gather data for a specified company. However, direct access to news within a precise date range from the website is not available. Nonetheless, you can choose a time frame such as ``within a year'', ``within a month'', or ``within a week'' for your news search. Consequently, only the streaming interface is supported.

\textbf{Seeking Alpha} is a premier online platform that provides investors, analysts, and financial enthusiasts with a wealth of valuable information, analysis, and insights on global financial markets. Launched in 2004, Seeking Alpha has become a trusted destination for individuals seeking intelligent investment ideas and staying informed about the latest market trends. Investors can exchange their ideas or their understanding of the market on that platform. There is also much useful information including news on the platform. We can gather news from a date range directly from the website and the general news and company news are both provided by the website. Consequently, this data source supports both streaming and data range interfaces.

\textbf{Penny Stock} is the top online destination for all things Micro-Cap Stocks. On Penny Stocks, one will find a comprehensive list of Penny Stocks \& discover the best Penny Stocks to buy, top penny stock news, and micro-cap stock articles. It provides a unique high-risk, high-reward investment opportunity and is happy to be there with its users every step of the way. We can gather data related to a certain company. However, we can only gather data in the streaming format.

\textbf{Market Watch} provides the latest stock market, financial and business news. One can get stock market quotes, personal finance advice, company news, etc. They offer all the latest stock market news and currencies market news. We can gather data related to a certain company. However, we can only gather data in the streaming format.

\textbf{Tip Ranks} is a financial analysis and research platform that allows users to track the performance and accuracy of financial analysts, hedge fund managers, and bloggers. With access to a database of over 10 million data points, TipRanks provides users with actionable insights and investment ideas. They strive to create a fair and equal environment by democratizing access to institutional research tools and data, making them available to everyone. We can gather data related to a certain company. However, we can only gather data in the streaming format.

\textbf{The Fly} is a leading digital publisher of real-time financial news. Their mission is to report and explain the news impacting publicly traded companies. They deliver rapid and up-to-the-minute coverage of breaking news pertaining to publicly traded companies. We can gather data related to a certain company. However, We can only gather data in the streaming format.

\begin{table}
   \caption{News data in FinGPT.}
   \label{tbl:3}
   \setlength{\tabcolsep}{4pt}
   \centering
    \begin{tabular}{ccccc}
        \toprule  
         Source Name  & Related Market & Source Type & Specific Company & Daily Pricing  \\
        \midrule
         Yahoo       &  US Stocks & Streaming & \ding{55} & Free \\
         Reuters     &  US Stocks & Streaming & \checkmark & Free  \\
         Seeking Alpha &  US Stocks &  Data Range / Streaming & \checkmark & Free  \\
         Penny Stocks &  US Stocks &  Streaming & \checkmark & Free  \\
         Market Watch &  US Stocks & Streaming  & \checkmark & Free   \\
         Tip Ranks    &  US Stocks & Streaming  & \checkmark & \$1$\sim$\$1.67  \\
         The Fly     &  US Stocks  & Streaming  & \checkmark  & Free  \\
         Talk Markets &  US Stocks & Streaming  & \checkmark & Free  \\
         Alliance News &  US Stocks & Streaming  & \checkmark & Free  \\
         Guru Focus  &  US Stocks  & Streaming  & \checkmark  & \$1.37$\sim$\$6.57   \\
         Investor Place &  US Stocks  & Streaming  & \checkmark & Free  \\
         FMP          &  US Stocks & Streaming & \checkmark & \$0.47$\sim$\$3.30  \\
         Sina        &  CN Stocks  & Data Range / Streaming & \checkmark  & Free  \\
         Eastmoney    &  CN Stocks  & Streaming & \checkmark  & Free \\
         Yicai       &  CN Stocks   & Streaming & \ding{55}  & Free  \\
         CCTV        &  CN Stocks  & Data Range / Streaming  &\ding{55}  & Free  \\
         Tushare     &  CN Stocks  & Data Range / Streaming  & \checkmark  & \$0.46  \\
         FinnHub     &  US Stocks  & Data Range / Streaming  & \checkmark & \$1.67$\sim$\$5  \\
         CNBC        &  US Stocks  & Streaming  & \checkmark  & Free \\

        \bottomrule
    \end{tabular}
\end{table}

\textbf{Talk Markets} is a financial content site that is truly customized, optimized, and socialized. They cover the entire breadth of diverse financial realms but are customized and tailored to each individual user. Their interests, preferences, and level of investment sophistication influence what content they see and in what medium. We can gather data related to a certain company. However, we can only gather data in the streaming format.

\textbf{Alliance News} provides real-time news coverage of the companies, markets, and economies that matter the most to investors globally. They report on the 500+ companies that make up the leading stock indices around the world, including the Stoxx Global 150, Dow 30, Nasdaq 100, FTSE 100, DAX 30, and CAC 401. Their journalists and partner news agencies track key data reports, central bank decisions, and government policy debates from the biggest and the most interconnected economies. We can gather data related to a certain company. However, we can only gather data in the streaming format.

\textbf{Guru Focus} is a financial news and research platform that focuses on what the stock market’s insiders and most well-known investors are trading. They track the trading action of over 175 ``gurus'' – typically fund managers and wealthy individual investors – and company CEOs and CFOs to help traders get an edge on the market. The service allows users to track the market, the gurus, and even institutional investors. We can gather data related to a certain company. However, we can only gather data in the streaming format.

\textbf{Investor Place} is an investing and financial news site that provides investors with free stock picks, options trades, market news, and actionable commentary. They provide millions of investors with insightful articles and stock market news. Their analysts offer research and advice to help investors make big gains from the world’s biggest macroeconomic and geopolitical events. We can gather data related to a certain company. However, We can only gather data in the streaming format.

\textbf{Financial modeling prep (FMP)} is a leading financial data and modeling platform that equips investors, analysts, and financial professionals with a wealth of robust tools and comprehensive data to make informed investment decisions. With its user-friendly interface and diverse range of features, FMP serves as a one-stop solution for individuals and businesses seeking reliable financial information. On the FMP platform, the news is provided in the streaming format, so that we can call the API to get the news data for a certain company from now for certain pages. The news cover almost all the mainstream stocks of the US market.

\textbf{Sina} is one of the biggest news websites in China and its financial news also covers a wide range of aspects. The content of the news is in Chinese so we may not only use them to fine-tune Chinese Models or Analyze them in Chinese but also fine-tune some bi-language models which may enhance model ability in cross languages. The Sina data source provides news from various aspects, not only news financial news but also news in politics, entertainment, sports, etc. Both streaming and data range interfaces are supported. However, most of the data from Sina is in streaming format and only the financial general news can be reached in the date range format.

\textbf{Eastmoney} is one of the biggest general financial platforms in China. Not only does it provide information like news or price data, but it also provides a forum for investors to exchange ideas. The platform provides both general news and news about certain companies, but we can only gather the data in the streaming format.

\textbf{Yicai} is also one of the most professional financial media in China. Although the quantity of the total news on that platform is not as much as Sina or Eastmoney, the news on that platform is written by professional financial critics or financial writers. We can only gather the data in the streaming format.

\textbf{CCTV} is the official media of China. Its everyday news can demonstrate the development of China directly. Besides, it is also one of the best ways for us to gather the important government policy and attitudes toward certain incidents. Since the Akshare platform has connected the CCTV data source and has covered the news since 1994, we directly call the API of Akshare in our program and the data can be accessed in both streaming and date range formats.

\textbf{Tushare} used to be one of the best financial data sources in China. Various types of data from price data to alternative data to statics data of the whole country can all be found on that platform. Although some of the key factors are charged by Tushare now, it is still affordable for most investors and researchers. Besides, there are some free data including news data provided by Tushare. To get full access to the news data, you might be charged 500 Chinese Yuan every year, which is equal to about 71 dollars. Both streaming and date range interfaces are supported.

\textbf{FinnHub} is a leading financial data platform that empowers investors, traders, and developers with access to a wide range of real-time and historical financial data. With its extensive coverage and user-friendly interface, Finnhub has become a go-to resource for individuals and businesses seeking reliable and accurate financial information. As for news information, Finnhub provides free news for a whole year and more news for charged plans. Since the market is highly dynamic, news within a year is enough for us to fine-tune models or analyze the market. Both streaming and date range interfaces are supported.

\begin{table}
   \caption{The total number of stock-related documents from some mainstream news data sources for six big tech companies  in the U.S. market. }
   \label{tbl:4}
   \centering
    \begin{tabular}{ccccccccc}
        \toprule 
         Source Name   & AAPL & AMZN & NFLX & GOOGL & MSFT & NVDA & Total \\
        \midrule 
         Yahoo          &  67525 &  164615 &  40515 & 129940 & 36500 & 27010 & 466105 \\
         Reuters        &  3837 & 3040 & 1413 & 5039 & 2423 & 750 & 16502 \\
         Seeking Alpha  &  9535 &  6706 &  2919 & 3606 & 6350 & 2104 & 31220 \\
         Penny Stocks   &  471 &  325 &  89 & 132 & 97 & 40 & 1154 \\
         Market Watch   &  51251 & 33010 & 13646 & 32842 & 30700 & 7700 & 169149 \\
         Talk Markets   &  4590 &  4950 &  1540 & 3400 & 2970 & 1320 & 18770 \\
         FMP            &  35026 & 33040 & 11284 & 22712 & 17323 & 10858 & 130243 \\
         
        \midrule
         Total & 174026  & 249401 & 72252 & 200182 & 97066 & 50264 & 843191 \\
        
        \bottomrule  
    \end{tabular}
\end{table}

\textbf{CNBC} is an American basic cable business news channel and website that provides business news programming on weekdays from 5:00 a.m. to 7:00 p.m., Eastern Time. They also broadcast talk shows, investigative reports, documentaries, infomercials, reality shows, and other programs at all other times. Their website provides the latest stock market, financial and business news We can gather data related to a certain company. However, We can only gather data in the streaming format.

To offer an understanding of the data volume, Table~\ref{tbl:4} presents a summary of the total count of stock-related documents obtained from prominent mainstream news sources.

\subsubsection{Social Media}

The social media data sources are summarized in Table~\ref{tbl:5}.

\begin{table}
   \caption{Social media data in FinGPT.}
   \label{tbl:5}
   \setlength{\tabcolsep}{4pt}
   \centering
    \begin{tabular}{ccccc}
        \toprule
         Source Name  & Related Market & Source Type & Specific Company & Daily Pricing \\
        \midrule
         Twitter  & US Stocks  & Date Range / Streaming & \checkmark & Free \\
         Reddit   & US Stocks &  Streaming & \checkmark & Free \\
         Weibo   & CN Stocks &  Date Range / Streaming & \checkmark & Free \\
         Xueqiu  & CN Stocks &  Streaming & \checkmark & Free \\
         Facebook & US Stocks  &  Streaming & \checkmark & Free \\
         StockTwits & US Stocks & Streaming & \checkmark & Free \\
         Eastmoney & CN Stocks & Streaming & \checkmark & Free \\

        \bottomrule   
    \end{tabular}
\end{table}

\textbf{Twitter} is a social media platform that serves the public conversation. It provides a free and safe space for people to talk and share information in real time. Users can join the conversation, follow accounts, see their Home Timeline, and catch up on Tweets from the people they know. Thanks to the powerful search function of Twitter, we can search for the specific company of interest. It also supports both streaming and data range interfaces.

\textbf{Reddit} is a social news aggregation, web content rating, and discussion website. It is a network of communities based on people’s interests where registered members submit content to the site such as links, text posts, and images, which are then voted up or down by other members. Posts are organized by subject into user-created boards called ``subreddits'', which cover a variety of topics including news, science, movies, video games, music, books, fitness, food, and image-sharing.

The subreddit ``wallstreetbets'' is a community on Reddit.com where users discuss stock and options trading. It has become notable for its profane nature, aggressive trading strategies, and role in the GameStop short squeeze that caused losses on short positions in U.S. firms topping \$70 billion in a few days in early 2021. We can not only gather data related to certain companies, but we can also gather the market changes through subreddits like "wallstreetbets". Due to the limits of the platform, only the streaming format is supported.

\textbf{Weibo} is a Chinese microblogging website launched by Sina Corporation on August 14, 2009. It is one of the biggest social media platforms in China, with over 500 million registered users. Users can create and post short messages, known as ``weibo'', and share them with their followers. Weibo also allows users to share multimedia content such as photos and videos. Thanks to the platform, we are able to search for any keyword we want, and if we want to gather the data for a certain date range, we just need to log in to that platform by passing cookies. Thus, it supports both streaming and data range interfaces.

\textbf{Xueqiu} is a Chinese social network platform for investors. It provides a space for users to share their insights and opinions on financial markets, stocks, and other investment opportunities. The platform also offers real-time quotes, professional data analysis, and a variety of investment tools to help users make informed decisions. We can gather data related to a certain company. However, we can only gather data in the streaming format.

\textbf{Facebook} is a social networking website that allows users to connect with friends and family, and share content with others. It provides a platform for users to create a personal profile, add other users as friends, and exchange messages, including automatic notifications when they update their profile. Additionally, users may join common-interest user groups, organized by workplace, school or college, or other characteristics. We can use the search function to search for tweets related to certain companies. However, we can only gather data in the streaming format.

\textbf{StockTwits} is a social media platform designed for sharing ideas between investors, traders, and entrepreneurs. The platform allows users to create a personalized financial news feed by following their favorite stocks, assets, and other users. With millions of investors, StockTwits is considered the voice of global finance. We can gather data related to a certain company. However, we can only gather data in the streaming format.

\textbf{Eastmoney} is a Chinese financial portal that provides professional financial news and data on stocks, markets, securities, funds, banking, insurance, trusts, futures, gold, and more. The website offers a wide range of tools and services for investors, including real-time quotes, data analysis, and investment advice. Eastmoney is a popular source of financial information for Chinese investors. We can gather data related to a certain company. However, we can only gather data in the streaming format.

\subsubsection{Filing}

The filing data sources are summarized in Table~\ref{tbl:6}.

\begin{table}
   \caption{Filings data in FinGPT.}
   \label{tbl:6}
   \setlength{\tabcolsep}{4pt}
   \centering
    \begin{tabular}{ccccc}
        \toprule
        Source Name  & Related Market & Source Type & Specific Company & Daily Pricing \\
        \midrule 
        SEC   & US Market & Date Range / Streaming & \checkmark & Free \\
        Juchao & CN Market & Date Range / Streaming & \checkmark & Free \\

        \bottomrule  
    \end{tabular}
\end{table}

\textbf{SEC} is the official website of the U.S. Securities and Exchange Commission (SEC), an independent federal government agency responsible for protecting investors, maintaining fair and orderly functioning of securities markets, and facilitating capital formation. The website provides a wealth of information and resources for investors, including news, alerts, and educational materials. It also allows users to access and search SEC filings and forms electronically through the EDGAR system. Thanks to the powerful search function of SEC, we can search for the company we want. It supports both streaming and data range interfaces.

\textbf{Juchao} is a designated information disclosure platform for companies listed on the Shenzhen Stock Exchange. The website provides a wealth of information and resources for investors, including company announcements, financial reports, and market data. It also allows users to access and search for information about listed companies and their securities. Thanks to the powerful search function of Juchao, we can not only search for the company we want. It supports both streaming and data range interfaces.

\subsubsection{Research Dataset}

The research datasets are summarized in Table~\ref{tbl:7}.

\textbf{Stocknet}~\cite{xu-cohen-2018-stock} dataset is a comprehensive dataset for stock movement prediction from tweets and historical stock prices 1. It consists of two-year price movements from 01/01/2014 to 01/01/2016 of 88 stocks 1. These stocks come from all 8 stocks in the Conglomerates sector and the top 10 stocks in capital size in each of the other 8 sectors.

\textbf{CHRNN}~\cite{wu2018hybrid} dataset is associated with a proposed model called CHRNN, which stands for Hybrid Deep Sequential Modeling for Social Text-Driven Stock Prediction. The CHRNN model and dataset aim to provide a solution for social text-driven stock prediction. This paper was accepted by CIKM’18.

\textbf{The TradeTheEvent (TTE)}~\cite{zhou-etal-2021-trade} dataset is an open-source dataset for corporate event detection and news-based stock prediction benchmark. It is released by Zhihan Zhou, Liqian Ma, and Han Liu as part of their paper “Trade the Event: Corporate Events Detection for News-Based Event-Driven Trading” published in Findings of ACL 2021. 

\textbf{Astock}~\cite{zou-etal-2022-astock} is an open-source dataset and automated stock trading system based on stock-specific news analyzing model 1. It was developed by Jinan Zou and introduced in a paper accepted by FinNLP 2022 from IJCAI. The dataset and code are available on GitHub.

\textbf{FPB~\cite{malo2014good}} dataset entails a sentiment classification task on sentences from financial news. The labels for classification are ``neutral'', ``positive'', and ``negative''.

\textbf{FiQA SA~\cite{maia201818}} is to forecast sentiment in English financial news and microblog headlines, which were originally released as part of the 2018 challenge on financial question answering and opinion mining.

 \begin{table}
   \caption{Research data in FinGPT.}
   \label{tbl:7}
   \centering
    \begin{tabular}{cccccc}
        \toprule    
        Source Name & Specific Company & Source Type \\
        \midrule
        Stocknet & \checkmark & Social Media \\
        CHRNN  & \checkmark & Social Media \\
        TTE & \ding{55} & News \\
        Astock & \checkmark & News  \\
        FiQA SA & \ding{55} & News \& Social Media \\
        FPB & \ding{55} &  News  \\

        \bottomrule 
    \end{tabular}
\end{table}

\subsection{Example Codes for Accessing Data}
\label{sec:C2}

We offer API examples that demonstrate how to access various data sources. You can find more examples at \url{https://github.com/AI4Finance-Foundation/FinNLP}.

\subsubsection{News}

\textbf{CNBC}

\begin{python}
from finnlp.data_sources.news.cnbc_streaming import CNBC_Streaming

news_downloader = CNBC_Streaming()
news_downloader.download_streaming_search(keyword = "apple", rounds = 3)
\end{python}

\begin{CJK}{UTF8}{gbsn}
\textbf{Yicai / 第一财经}

\begin{python}
from finnlp.data_sources.news.yicai_streaming import Yicai_Streaming

news_downloader = Yicai_Streaming()
news_downloader.download_streaming_search(keyword = keyword, rounds = 3)
\end{python}
where \pyth{keyword} is a Simplified Chinese phrase like ``茅台''.
\end{CJK}

\textbf{Investor Place}

\begin{python}
from finnlp.data_sources.news.investorplace_streaming import InvestorPlace_Streaming

news_downloader = InvestorPlace_Streaming()
news_downloader.download_streaming_search(keyword = "apple", rounds = 3)
\end{python}

\textbf{Guru Focus}

\begin{python}
from finnlp.data_sources.news.gurufocus_streaming import GuruFocus_Streaming

news_downloader = GuruFocus_Streaming()
news_downloader.download_streaming_search(keyword = "AAPL", rounds = 3)
\end{python}

\textbf{Alliance News}

\begin{python}
from finnlp.data_sources.news.alliancenews_streaming import AllianceNews_Streaming

news_downloader = AllianceNews_Streaming()
news_downloader.download_streaming_search(rounds = 3)
\end{python}

\textbf{Talk Market}

\begin{python}
from finnlp.data_sources.news.talkmarkets_streaming import TalkMarkets_Streaming

news_downloader = TalkMarkets_Streaming()
news_downloader.download_streaming_search(keyword = "apple", rounds = 3)
\end{python}

\textbf{The Fly}

\begin{python}
from finnlp.data_sources.news.thefly_streaming import TheFly_Streaming

news_downloader = TheFly_Streaming()
news_downloader.download_streaming_search(keyword = "AAPL", rounds = 3)
\end{python}

\textbf{Tip Rank}

\begin{python}
from finnlp.data_sources.news.tipranks_streaming import TipRanks_Streaming

news_downloader = TipRanks_Streaming()
news_downloader.download_streaming_search(keyword = "apple", rounds = 3)
\end{python}

\textbf{Market Watch (Date Range)}

\begin{python}
from finnlp.data_sources.news.marketwatch_date_range import MarketWatch_Date_Range

start_date = "2022-06-01"
end_date = "2022-06-30"
keyword = "apple"

news_downloader = MarketWatch_Date_Range()
news_downloader.download_date_range_search(keyword = "apple", start_date = start_date, end_date = end_date)
\end{python}

\textbf{Market Watch (Streaming)}

\begin{python}
from finnlp.data_sources.news.marketwatch_streaming import MarketWatch_Streaming

news_downloader = MarketWatch_Streaming()
news_downloader.download_streaming_search(keyword = "apple", rounds = 3)
\end{python}

\textbf{Penny Stock}

\begin{python}
from finnlp.data_sources.news.pennystocks_streaming import PennyStocks_Streaming

news_downloader = PennyStocks_Streaming()
news_downloader.download_streaming_search(keyword = "apple", rounds = 3)
\end{python}

\textbf{Seeking Alpha}

\begin{python}
from finnlp.data_sources.news.seekingalpha_date_range import SeekingAlpha_Date_Range

start_date = "2023-06-01"
end_date = "2023-06-30"
stock = "AAPL"

news_downloader = SeekingAlpha_Date_Range()
news_downloader.download_date_range_stock(start_date, end_date, stock)
\end{python}

\textbf{Reuters}

\begin{python}
from finnlp.data_sources.news.reuters_streaming import Reuters_Streaming

news_downloader = Reuters_Streaming()
news_downloader.download_streaming_search(keyword = "apple", rounds = 3)
\end{python}

\textbf{Sina Finance}

\begin{python}
from finnlp.data_sources.news.sina_finance_date_range import Sina_Finance_Date_Range

start_date = "2016-01-01"
end_date = "2016-01-01"
config = {
    "use_proxy": "china_free",
    "max_retry": 5,
    "proxy_pages": 5,
}

news_downloader = Sina_Finance_Date_Range(config)
news_downloader.download_date_range_all(start_date,end_date)
news_downloader.gather_content()
\end{python}

\textbf{Eastmoney}

\begin{python}
from finnlp.data_sources.news.eastmoney_streaming import Eastmoney_Streaming

pages = 3
stock = "600519"
config = {
    "use_proxy": "china_free",
    "max_retry": 5,
    "proxy_pages": 5,
}

news_downloader = Eastmoney_Streaming(config)
news_downloader.download_streaming_stock(stock,pages)
\end{python}

\textbf{Finnhub / Yahoo}

\begin{python}
from finnlp.data_sources.news.finnhub_date_range import Finnhub_Date_Range

start_date = "2023-01-01"
end_date = "2023-01-03"
config = {
    "use_proxy": "us_free",
    "max_retry": 5,
    "proxy_pages": 5,
    "token": "YOUR_FINNHUB_TOKEN"  # Avaliable at https://finnhub.io/dashboard
}

news_downloader = Finnhub_Date_Range(config)
news_downloader.download_date_range_stock(start_date,end_date)
news_downloader.gather_content()
\end{python}

\subsubsection{Social Media}

\textbf{Eastmoney}

\begin{python}
from finnlp.data_sources.social_media.eastmoney_streaming import Eastmoney_Streaming

pages = 3
stock = "600519"

downloader = Eastmoney_Streaming()
downloader.download_streaming_stock(stock, pages)
\end{python}

\textbf{Facebook}

\begin{python}
from selenium import webdriver
import json
from finnlp.data_sources.social_media.facebook_streaming import Facebook_Streaming

# Get cookies
browser = webdriver.ChromiumEdge()
browser.get('https://www.facebook.com')
cookies = browser.get_cookies() 
with open("cookies.json", "w", encoding="utf-8") as cks:
    json.dump(cookies, cks)

# load cookies
with open("cookies.json", "r", encoding="utf-8") as cks: 
    cookies = json.load(cks)

config = {
    "cookies":cookies, 
    "headless": False,
    "stealth_path":"../../FinNLP/finnlp/data_sources/social_media/stealth.min.js"
    }
pages = 3
stock = "AAPL"

downloader = Facebook_Streaming(config)
downloader.download_streaming_stock(stock, pages)
\end{python}

\begin{CJK}{UTF8}{gbsn}
\textbf{Xueqiu / 雪球}

\begin{python}
from finnlp.data_sources.social_media.xueqiu_streaming import Xueqiu_Streaming

pages = 3
downloader = Xueqiu_Streaming()
downloader.download_streaming_stock(stock, pages)
\end{python}
where \pyth{stock} is a Simplified Chinese phrase like ``茅台''.
\end{CJK}

\textbf{Stocktwits Streaming}

\begin{python}
from finnlp.data_sources.social_media.stocktwits_streaming import Stocktwits_Streaming

pages = 3
stock = "AAPL"
config = {
    "use_proxy": "us_free",
    "max_retry": 5,
    "proxy_pages": 2,
}

downloader = Stocktwits_Streaming(config)
downloader.download_streaming_stock(stock, pages)
\end{python}

\textbf{Reddit Wallstreetbets Streaming}

\begin{python}
from finnlp.data_sources.social_media.reddit_streaming import Reddit_Streaming

pages = 3
config = {
    # "use_proxy": "us_free",
    "max_retry": 5,
    "proxy_pages": 2,
}

downloader = Reddit_Streaming(config)
downloader.download_streaming_all(pages)
\end{python}

\begin{CJK}{UTF8}{gbsn}
\textbf{Weibo Date Range}

\begin{python}
from finnlp.data_sources.social_media.weibo_date_range import Weibo_Date_Range

start_date = "2016-01-01"
end_date = "2016-01-02"
config = {
    "use_proxy": "china_free",
    "max_retry": 5,
    "proxy_pages": 5,
    "cookies": "Your_Login_Cookies",
}

downloader = Weibo_Date_Range(config)
downloader.download_date_range_stock(start_date, end_date, stock = stock)
\end{python}
where \pyth{stock} is a Simplified Chinese phrase like ``茅台''.
\end{CJK}

\begin{CJK}{UTF8}{gbsn}
\textbf{Weibo Streaming}

\begin{python}
from finnlp.data_sources.social_media.weibo_streaming import Weibo_Streaming

rounds = 3
config = {
    "use_proxy": "china_free",
    "max_retry": 5,
    "proxy_pages": 5,
    "cookies": "Your_Login_Cookies",
}

downloader = Weibo_Streaming(config)
downloader.download_streaming_stock(stock = stock, rounds = rounds)
\end{python}
where \pyth{stock} is a Simplified Chinese phrase like ``茅台''.
\end{CJK}

\subsubsection{Filing}

\textbf{SEC}

\begin{python}
from finnlp.data_sources.company_announcement.sec import SEC_Announcement

start_date = "2020-01-01"
end_date = "2020-06-01"
stock = "AAPL"
config = {
    "use_proxy": "us_free",
    "max_retry": 5,
    "proxy_pages": 3,
}

downloader = SEC_Announcement(config)
downloader.download_date_range_stock(start_date, end_date, stock = stock)
\end{python}

\textbf{Juchao}

\begin{python}
from finnlp.data_sources.company_announcement.juchao import Juchao_Announcement

start_date = "2020-01-01"
end_date = "2020-06-01"
stock = "000001"
config = {
    "use_proxy": "china_free",
    "max_retry": 5,
    "proxy_pages": 3,
}

downloader = Juchao_Announcement(config)
downloader.download_date_range_stock(start_date, end_date, stock = stock, get_content = True, delate_pdf = True)
\end{python}

\subsection{Challenges and Solutions}

In this subsection, we discuss the challenges that we have encountered when obtaining the data and the solutions.





\subsubsection{Asynchronous Crawling}

There is a large amount of content to be crawled, so using asynchronous multi-threading/processes can improve efficiency. To implement this, we used the built-in Python \textsf{multiprocessing} library was used for asynchronous multi-threading pooling. The general code structure is as follows:

\begin{python}
import multiprocessing as mp
 import os
 
 company_list = [...]
 date_list = [...]
 
 pool = mp.Pool(40)
 pool_list = []
 for stock in company_list:
     path = f"./results/{stock}"
     os.makedirs(path, exist_ok=True)
     date_list_stock = os.listdir(path) 
     for date in date_list:
         if not f"{date}.csv" in date_list_stock:
             res = pool.apply_async(download_news, args=((stock, date),),error_callback = lambda x:print(x))
             pool_list.append(res)
 
 pool.close()
 pool.join()
\end{python}

\subsubsection{Incremental Saving}

There is a possibility of the crawler encountering failures, so we need to save successful results while also being able to re-crawl content that failed. In our implementation, we save the results of each stock separately for each day, allowing for incremental crawling. This also enables better handling of crawl failures. Specifically,an example file structure is as follows:

\begin{lstlisting}
    - AAPL
        - 2023-01-01.csv
        - 2023-01-02.csv
        - 2023-01-03.csv
        ...
    - ABNB
        - 2023-01-01.csv
        - 2023-01-02.csv
        ...
    ...
\end{lstlisting}



\section{Quantitative Analysis of Training Cost}
\label{sec:D}

In this section, we present an analysis of the training costs for FinGPT, BloombergGPT, and other LLMs. It should be noted that our aim is to provide a general idea of the training costs, relying on AWS pricing as a reference. The precise expenditure can be challenging to ascertain, as it might vary significantly depending on the distinct training systems employed by different parties.

\textbf{Cost per GPU hour}. For A100 GPUs, the AWS p4d.24xlarge instance, equipped with 8 A100 GPUs, is used as a benchmark to estimate the costs. Note that BloombergGPT~\cite{wu2023bloomberggpt} also used p4d.24xlarge. As of July 11, 2023, the hourly rate for this instance stands at \$32.77\footnote{\url{https://aws.amazon.com/ec2/instance-types/p4/}}. Consequently, the estimated cost per GPU hour comes to \$32.77 divided by 8, resulting in approximately \$4.10. With this value as the reference unit price (i.e., $1$ GPU hour), we proceed to compute the training cost for each LLM based on the number of GPU hours consumed. For V100 GPUs, we use AWS p3dn.24xlarge. As of July 11, 2023, the hourly rate for this instance stands at \$31.218\footnote{\url{https://aws.amazon.com/ec2/instance-types/p3/}}. The estimated cost per GPU hour is \$3.90. 

Training costs of LLMs: 
\begin{itemize}[leftmargin=*]
    \item \textbf{FinGPT:} Our LoRA-based fine-tuning process was conducted on a DGX-2 server with 8 A100 GPUs over a duration of 8 hours. Hence, there were in total $8 \times 8 = 64$ GPU hours. By multiplying this by the cost per GPU hour, the total fine-tuning cost was $64 \times \$4.10 = \$262.40$, which is approximately \$262.
    \item \textbf{BloomberGPT~\cite{wu2023bloomberggpt}:} The training process engaged 512 A100 GPUs for approximately 53 days. This translates to $512 \times 53 \times 24 = 651,264$ GPU hours. By multiplying this number by the per GPU hour rate, $651,264 \times \$4.10 = \$2,670,182.40$, the total estimated cost of the training process amounts to approximately \$2.67 million.
    \item \textbf{ChatGLM2~\cite{du2021glm}:} The training process engaged 64 V100 GPUs for approximately 2.5 days. This translates to $64 \times 2.5 \times 24 = 3,840$ GPU hours. 
    By multiplying this number by the per GPU hour rate, $3,840 \times \$3.90 = \$14,976.00$.
    \item \textbf{GPT-NeoX~\cite{black2022gpt}:} The training was conducted using 12 servers, each equipped with 8 A100 GPUs, so 96 A100 GPUs in total. The training process took approximately $1,830$ hours. Consequently, there were in total $96 \times 1,830 = 175,680$ GPU hours. Multiplying this figure by our per GPU hour rate, the resulting cost is $175,680 \times \$4.10 = \$720,288.00$, which is roughly \$0.72 million.
    \item \textbf{BLOOM~\cite{scao2022bloom}:} The model underwent training using 384 A100 GPUs over a duration of 105 days. This translates to a total of $384 \times 105 \times 24 = 967,680$ GPU hours. By multiplying this number by the cost per GPU hour, the total training expense amounts to $967,680 \times \$4.10 = \$3,967,488.00$, which is approximately \$3.97 million.
     \item \textbf{LLaMA~\cite{scao2022bloom}:} The training process utilized 2048 A100 GPUs over a period of 21 days. Consequently, the model's training took $2048 \times 21 \times 24 = 1,032,192$ GPU hours. Multiplying this figure by our per GPU hour rate, the approximate training cost is $1,032,192 \times \$4.10 = \$4,231,987.20$, or around \$4.23 million.
    
\end{itemize}

\section{Performance Metrics}
\label{sec:E}

We describe how metrics used in the experiments are calculated, including accuracy, F1 score, and average cumulated return rate (CRR).

\subsection{Accuracy}

The accuracy metric is used for classification problems. It is calculated as the number of correct predictions divided by the total number of input samples,
\begin{equation*}
    \text{Accuracy} = \frac{\text{True Positives + True Negatives}}{\text{True Positives + True Negatives + False Positives + False Negatives}},
\end{equation*}
where True Positives (TP) is the number of true positives i.e., the number of positive cases that are correctly classified as positive, True Negatives (TN) is the number of true negatives i.e., the number of negative cases that are correctly classified as negative, False Positives (FP) is the number of false positives i.e., the number of negative cases that are incorrectly classified as positive, False Negatives (FN) is the number of false negatives i.e., the number of positive cases that are incorrectly classified as negative.

\subsection{F1 Score}
The F1 score is defined as the harmonic mean of the precision and recall. Precision is the number of true positive results divided by the number of all positive results, and recall is the number of true positive results divided by the number of positive results that should have been returned. Specifically,
\begin{equation*}
    \text{F1 Score} = \frac{2 * (\text{Precision * Recall})} {\text{Precision + Recall}}, 
\end{equation*}
where Precision = TP / (TP + FP), and Recall = TP / (TP + FN).

\subsection{Cumulative Return Rate (CRR)}

In the context of a simulated trading experiment, the Cumulative Return Rate can be calculated by dividing the final portfolio value by the initial portfolio value for each stock, subtracting $1$, and then multiplying by $100$ to get a percentage. For the $i$-th stock, we have
\begin{equation*}
    \text{CRR}_i = \left(\frac{\text{Final Portfolio Value}_i}{\text{Initial Portfolio Value}_i} - 1 \right) * 100\%,
\end{equation*}
where $\text{Final Portfolio Value}_i$ and $\text{Initial Portfolio Value}_i$ are the final and initial porfolio values for the $i$-th stock, respectively.

Then, the average CRR over $N$ stocks can be calculated as: 
\begin{equation*}
    \text{Average CRR} = \left( \frac{ \sum\limits_{i=1}^{N} \text{Final Portfolio Value}_i}{ \sum\limits_{i=1}^{N} \text{Initial Portfolio Value}_i} - 1 \right) * 100\%.
\end{equation*}
 It is important to note that the calculation of the final portfolio value would depend on the specific trading strategy employed, which includes the number of stocks bought or sold at each trade, the price at each trade, and other factors.

\section{Details of Labeling by Market Experiment}
\label{sec:F}

\subsection{Data}

The data are collected from the FMP data source, and in order to make market-related financial labels we collect the price data from yahoo finance. There are $582,734$ pieces of news in total. Then the data was split into train-valid and test sets. The split date was set to “2021-11-01". There are about 80\% (465,645) of the data was before or equal to the split date and the rest are later. We used the first part as the train\&valid set and the latter part as the test set. So there are $465,645$ pieces of news in the train set and 117,089 pieces of news in the test set. Then we split the train\&valid set to train set and valid set randomly with the proportion of 0.8 to 0.2 and the random seed was set to 42.

\subsection{Market-Related Financial Labels}

The use of labeled data is crucial when fine-tuning Language Models (LLMs). However, labeling data can be expensive, and the assigned labels may have limited relevance to the market. Instead, we use an alternative approach that leverages market behavior for labeling. Specifically, we apply thresholds based on changes in stock prices to categorize company-related news into three labels: ``Positive'', ``Negative'', and ``Neutral''. In this experiment, a threshold of 2\% was set. Accordingly, if the percentage change in stock price exceeds 2\%, the related news is labeled as ``Positive''. Conversely, if the percentage change is below -2\%, the news is labeled as ``Negative''. For changes between -2\% and 2\%, the news is labeled as ``Neutral''. After applying these labels, approximately 37\% of the news is categorized as ``Negative'', around 42\% as ``Positive'', and roughly 21\% as ``Neutral''.

\subsection{Training Details}

For the LoRA setting, the LoRA rank was 8, the LoRA Alpha was 16 and the dropout rate of the LoRA linear function was set to 0.05. And for other parameters, the batch size was 128 and the learning rate was set to 3e-3. For more details, please refer to our website.

\section{Details of Supervised Fine-tuning Experiment}
\label{sec:G}



\subsection{Datasets and Splits}

The datasets and splits follow the setting described in BloombergGPT~\cite{wu2023bloomberggpt}.

\subsubsection{Financial Phrasebank (FPB)}

FPB dataset comprises a sentiment classification task involving approximately 5,000 sentences in English extracted from financial news concerning companies listed on OMX Helsinki. The sentiment annotations, categorized as positive, negative, or neutral, are determined from the perspective of an investor. Any news that could potentially benefit or harm an investor is considered positive or negative, respectively, while news that does not have such an impact is labeled as neutral. Each sentence is annotated by 5 to 8 annotators who possess sufficient knowledge of finance, while the source sentences themselves are written by financial journalists. To illustrate, news discussing a decline in revenue would be assigned a negative label, whereas news highlighting company growth would be labeled as positive. Various configurations of this dataset exist, each denoting the percentage agreement among annotators (e.g., $\ge$50\%, $\ge$66\%, $\ge$75\%, 100\%). For our purposes, we have chosen to utilize the configuration with a minimum agreement threshold of 50\%.

As there is no official train-test split available, we tried to follow the split of BloombergGPT, randomly dividing the data into training and test sets by setting the seed to 42. The training split consists of 3,876 sentences, comprising 1,086 positive, 488 negative, and 2,302 neutral sentences, while the test set contains 970 sentences, including 277 positive, 116 negative, and 577 neutral sentences. The numbers of each sentiment in both the train set and test set are equal to Bloomberg's split, so we can assume that we use the same split. We have chosen to evaluate the performance using 5-shot experiments and report the F1 score weighted by support.

\subsubsection{FiQA SA}

FiQA SA aimed to predict aspect-specific sentiment in English financial news and microblog headlines. This task was included as part of the 2018 challenge on financial question answering and opinion mining. The original task involved annotating sentiment on a continuous scale ranging from -1 to +1, although specific details regarding the annotation process are not readily available. To adapt this regression dataset for a few-shot Language Model (LLM) setup, we converted it into a classification task with three categories: Negative (-1 $\le$ x < -0.1), neutral (-0.1 $\le$ x < +0.1), and positive (+0.1 $\le$ x $\le$ +1), where x represents the original sentiment score. This discretization approach was chosen following a manual examination of the dataset, similar to our approach with FPB.

For our experimental setup, we created a random split combining both microblogs and news. After the discretization process, our training set consisted of 970 sentences, with 572 positive, 300 negative, and 98 neutral sentences. Additionally, our test set comprised 243 sentences, with 146 positive, 79 negative, and 18 neutral sentences. We selected a five-shot setup and report the weighted F1 score.

\subsubsection{TFNS}

The Twitter Financial News Sentiment (TFNS)~\cite{twitter2022finance} dataset is an English-language compilation of finance-related tweets, meticulously annotated. Designed for sentiment analysis, this dataset encompasses 11,932 documents categorized with three distinct labels: ``Bearish'' (indicative of a negative sentiment), ``Bullish'' (signifying a positive sentiment), and ``Neutral''.

\subsubsection{NWGI}

The News With GPT Instruction (NWGI) dataset uses labels produced by ChatGPT. With a training set encompassing 16.2k samples and a test set comprising 4.05k samples, it offers not just seven classification labels — ranging from ``strong negative'' to ``strong positive'', but also provides a rationale for each label. This additional insight could prove invaluable for fine-tuning instructional approaches.

\subsection{Models}
We provide a detailed description of each model as follows.
\begin{itemize}
    \item \textbf{(Pre-trained) BloombergGPT:} A pre-trained FinLLM by Bloomberg~\cite{wu2023bloomberggpt}. Since the model is not open-sourced, we report the results provided in~\cite{wu2023bloomberggpt}.
    \item \textbf{(Pre-trained) ChatGLM2:} A pre-trained LLM in~\cite{du2021glm}.
    \item \textbf{(Pre-trained) Llama2:} A pre-trained LLM in~\cite{llama2}.
    \item \textbf{(Pre-trained) ChatGPT:} A pre-trained LLM by OpenAI. We use the provided OpenAI's API for evaluation.
    \item \textbf{(Pre-trained) GPT-4:} A pre-trained LLM by OpenAI~\cite{OpenAI2023GPT4TR}. We use the provided OpenAI's API for evaluation.
    \item \textbf{(Fine-tuned) ChatGPT:} A fine-tuned model with the data collected in FinGPT by using OpenAI's fine-tuning API.
    \item \textbf{(Fine-tuned) Llama2:} A fine-tuned model based on Llama2 with the data collected in FinGPT. It uses LoRA~\cite{hulora} fine-tuning techniques with fp16 parameter format.
    \item \textbf{(Fine-tuned) ChatGLM2:} A fine-tuned model based on ChatGLM2 with the data collected in FinGPT. It uses LoRA~\cite{hulora} fine-tuning techniques with fp16 parameter format.
    \item \textbf{(Fine-tuned) ChatGLM2 (8-bit):} A fine-tuned model based on ChatGLM2 with the data collected in FinGPT. It uses the int-8~\cite{dettmers2022llmint8} and LoRA~\cite{hulora} fine-tuning techniques.
    \item \textbf{(Fine-tuned) ChatGLM2 (QLoRA):} A fine-tuned model based on ChatGLM2 with the data collected in FinGPT. It uses the QLoRA~\cite{dettmers2023qlora} fine-tuning techniques.
\end{itemize}

\subsection{Training Details}

We fine-tuned our FinGPT model with the LoRA method on a server with an A100 GPU. For the LoRA setting, the LoRA rank was 8, the LoRA Alpha was 16 and the dropout rate of the LoRA linear function was set to 0.05. And for other parameters, the batch size was 16, and the learning rate was set to 1e-4. For more details, please refer to our website. 

\section{Example of Building a Factor Library}

\label{sec:H}

\textbf{Init Prompt}: You are now a senior software development engineer in top financial company. You will help me to build a factor library in Python today.

\textbf{Init Response}: \textsf{Certainly! I can help you with building a factor library in Python for financial analysis. A factor library typically consists of functions or classes that calculate various quantitative factors used in financial analysis, such as price-to-earnings ratio, moving averages, volatility measures, and so on. These factors can be used for stock selection, risk management, and portfolio optimization.}

\textsf{Here's an example of how you can start building a factor library in Python:}

\begin{python}
import pandas as pd
import numpy as np

def calculate_price_to_earnings_ratio(price, earnings):
    """
    Calculates the price-to-earnings (P/E) ratio given the price and earnings.
    """
    return price / earnings

def calculate_moving_average(data, window):
    """
    Calculates the simple moving average given a data series and a window size.
    """
    return data.rolling(window=window).mean()

def calculate_volatility(data, window):
    """
    Calculates the volatility given a data series and a window size.
    """
    return data.rolling(window=window).std()

# Usage example
stock_prices = pd.Series([100, 110, 105, 95, 115, 120, 125, 130, 125, 135])
pe_ratio = calculate_price_to_earnings_ratio(stock_prices, 10)
moving_avg = calculate_moving_average(stock_prices, 3)
volatility = calculate_volatility(stock_prices, 5)

print("P/E Ratio:", pe_ratio)
print("Moving Average:", moving_avg)
print("Volatility:", volatility)
\end{python}

\textsf{In this example, we have defined three functions: `calculate price to earnings ratio`, `calculate moving average`, and `calculate volatility`. Each function takes the necessary inputs and returns the calculated factor.}

\textsf{You can expand on this library by adding more functions for other factors you want to calculate. You can also consider creating classes to encapsulate related factors and provide additional functionalities.}

\textsf{Remember to import necessary libraries like `pandas` and `numpy` to handle data and perform calculations efficiently.}

\textsf{I hope this helps you get started with building a factor library in Python. Let me know if you have any further questions!}

\textbf{Prompt 1}: Please help me design 5 more fundamental factors.

\textbf{Response 1}: \textsf{Certainly! Here are five more fundamental factors commonly used in financial analysis:}

\textsf{1. Return on Equity (ROE): ROE measures a company's profitability by calculating the return generated on shareholders' equity. It is typically calculated as the ratio of net income to average shareholders' equity over a period of time.}
\begin{python}
def calculate_return_on_equity(net_income, shareholders_equity):
    """
    Calculates the return on equity (ROE) given the net income and shareholders' equity.
    """
    average_equity = (shareholders_equity.shift(1) + shareholders_equity) / 2
    return net_income / average_equity
\end{python}

\textsf{2. Debt-to-Equity Ratio (D/E): The D/E ratio assesses a company's leverage by comparing its total debt to shareholders' equity. A higher D/E ratio indicates higher financial risk and dependency on debt financing.}
\begin{python}
def calculate_debt_to_equity_ratio(total_debt, shareholders_equity):
    """
    Calculates the debt-to-equity (D/E) ratio given the total debt and shareholders' equity.
    """
    return total_debt / shareholders_equity
\end{python}

\textsf{3. Earnings per Share (EPS): EPS measures the portion of a company's profit allocated to each outstanding share of common stock. It is commonly used to assess a company's profitability on a per-share basis.}

\begin{python}
def calculate_earnings_per_share(net_income, average_outstanding_shares):
    """
    Calculates the earnings per share (EPS) given the net income and average outstanding shares.
    """
    return net_income / average_outstanding_shares
\end{python}
    
\textsf{4. Current Ratio: The current ratio measures a company's liquidity and ability to cover short-term liabilities. It is calculated by dividing current assets by current liabilities. A ratio greater than 1 indicates the company has more assets than liabilities.}

\begin{python}
def calculate_current_ratio(current_assets, current_liabilities):
    """
    Calculates the current ratio given the current assets and current liabilities.
    """
    return current_assets / current_liabilities
\end{python}

\textsf{5. Gross Margin: Gross margin represents the percentage of revenue that a company retains after deducting the cost of goods sold (COGS). It reflects a company's ability to generate profits from its core operations.}

\begin{python}
def calculate_gross_margin(revenue, cogs):
    """
    Calculates the gross margin given the revenue and cost of goods sold (COGS).
    """
    return (revenue - cogs) / revenue
\end{python}

\textsf{These are just examples, and you can customize them further based on your specific requirements or industry standards. Remember to provide the necessary inputs for each factor calculation, such as net income, total debt, current assets, etc., based on the financial data you have.}

\textsf{Feel free to modify or enhance these functions as needed. Let me know if you have any more questions!}

\section{Examples of Building Factors}
\label{sec:I}

\textbf{Init Prompt}: Now, you are a senior professional stock factor researcher in top financial companies. You will help me to build some factors in Python today.

\textbf{Init Response}: \textsf{Certainly! I'd be happy to help you with building stock factors in Python. Stock factors are quantitative variables used in financial analysis to capture different aspects of a company's performance or market dynamics. These factors are often used in quantitative investment strategies to make data-driven investment decisions.}

\textsf{To get started, could you please let me know what specific factors you are interested in building? Factors can range from fundamental metrics such as earnings and revenue growth to technical indicators like moving averages and relative strength indices. The more specific you can be about the factors you want to create, the better I can assist you.}

\textbf{Prompt 1}: Let's look at the data, it's stored in the CSV format and has the following columns: open, high, low, close, volume, and amount.

\textbf{Response 1}: \textsf{Great! We can start by loading the data from the CSV file and examining its structure. Python provides several libraries that make it easy to work with CSV files and perform data analysis. One popular library for data manipulation and analysis is pandas.}

\textsf{To begin, you'll need to install the pandas library if you haven't already. You can install it by running the following command:}

\begin{python}
pip install pandas
\end{python}

\textsf{Once you have pandas installed, you can use the following code to load the data from the CSV file:}

\begin{python}
import pandas as pd

# Load the data from the CSV file
data = pd.read_csv('your_file.csv')

# Display the first few rows of the data
print(data.head())
\end{python}

\textsf{Replace `your\_file.csv` with the actual path and filename of your CSV file. The `read\_csv` function in pandas reads the CSV file and stores it in a DataFrame, which is a tabular data structure. The `head` method displays the first few rows of the DataFrame.}

\textsf{After loading the data, we can proceed with building the stock factors. Please let me know which specific factors you would like to create using the columns you mentioned: open, high, low, close, volume, and amount.}

\textbf{Prompt 2}: OK. First, please show me a function to calculate the Fundamental Momentum Factor. Let's use FMF to indicate Fundamental Momentum Factor.

\textbf{Response 2}: \textsf{Certainly! The Fundamental Momentum Factor (FMF) is a factor that captures the momentum of a company's fundamental metrics over a specific time period. It can be calculated using different fundamental metrics such as earnings, revenue, or book value.}

\textsf{To calculate the FMF, we can follow these steps:}

\textsf{1. Select a fundamental metric, such as earnings per share (EPS), as the basis for calculating the FMF.}

\textsf{2. Determine a time period over which you want to calculate the momentum. For example, you might choose a 12-month period.}

\textsf{3. Calculate the percentage change in the chosen fundamental metric over the specified time period for each company.}

\textsf{4. Rank the companies based on the percentage change in the fundamental metric.}

\textsf{5. Standardize the ranked values to have a mean of 0 and a standard deviation of 1.}

\textsf{Here's a Python function that calculates the Fundamental Momentum Factor (FMF) based on the earnings per share (EPS) over a specified time period:}

\begin{python}
import pandas as pd

def calculate_fmf(data, metric='EPS', time_period=12):
    # Calculate the percentage change in the chosen metric over the specified time period
    data['metric_change'] = data.groupby('company')[metric].pct_change(time_period)
    
    # Rank the companies based on the percentage change in the metric
    data['rank'] = data.groupby('date')['metric_change'].rank(ascending=False)
    
    # Standardize the ranked values
    data['fmf'] = (data['rank'] - data['rank'].mean()) / data['rank'].std()
    
    return data['fmf']
\end{python}

\textsf{In the above function, `data` represents the DataFrame containing the stock data, where each row corresponds to a specific date and company. The `metric` parameter allows you to choose the fundamental metric you want to use (default is set to 'EPS'). The `time\_period` parameter indicates the number of periods (months in this case) over which you want to calculate the momentum (default is set to 12 months).}

\textsf{You can call this function by passing your data DataFrame, and it will return a Series containing the Fundamental Momentum Factor (FMF) values for each row of the data.}

\textsf{Please note that you'll need to have the necessary fundamental metric (EPS) data available in your dataset for this function to work correctly.}

\textbf{Prompt 3}: OK. First, please show me a function to calculate the Fundamental Momentum Factor. Let's use FMF to indicate Fundamental Momentum Factor.

\textbf{Response 3}: \textsf{Certainly! The Price Momentum Factor (PMF) captures the momentum of a stock's price movement over a specific time period. It can be calculated using various technical indicators such as moving averages or relative strength indices (RSI).}

\textsf{To calculate the PMF, we can follow these steps:}

\textsf{1. Select a technical indicator, such as the 12-month moving average, as the basis for calculating the PMF.}

\textsf{2. Determine a time period over which you want to calculate the momentum.}

\textsf{3. Calculate the percentage change in the chosen technical indicator over the specified time period for each stock.}

\textsf{4. Rank the stocks based on the percentage change in the technical indicator.}

\textsf{5. Standardize the ranked values to have a mean of 0 and a standard deviation of 1.}

\textsf{Here's a Python function that calculates the Price Momentum Factor (PMF) based on the 12-month moving average:}

\begin{python}
import pandas as pd

def calculate_pmf(data, indicator='close', time_period=12):
    # Calculate the percentage change in the chosen technical indicator over the specified time period
    data['indicator_change'] = data.groupby('company')[indicator].pct_change(time_period)
    
    # Rank the stocks based on the percentage change in the technical indicator
    data['rank'] = data.groupby('date')['indicator_change'].rank(ascending=False)
    
    # Standardize the ranked values
    data['pmf'] = (data['rank'] - data['rank'].mean()) / data['rank'].std()
    
    return data['pmf']
\end{python}

\textsf{In the above function, `data` represents the DataFrame containing the stock data, where each row corresponds to a specific date and company. The `indicator` parameter allows you to choose the technical indicator you want to use (default is set to 'close' for the closing price). The `time\_period` parameter indicates the number of periods (months in this case) over which you want to calculate the momentum (default is set to 12 months).}

\textsf{You can call this function by passing your data DataFrame, and it will return a Series containing the Price Momentum Factor (PMF) values for each row of the data.}

\textsf{Please note that you'll need to have the necessary price data available in your dataset for this function to work correctly. Also, keep in mind that there are various technical indicators you can use to calculate the PMF, so feel free to modify the function according to your specific requirements.}

\textbf{Prompt 4}: OK. First, please show me a function to calculate the Fundamental Momentum Factor. Let's use FMF to indicate Fundamental Momentum Factor.

\textbf{Response 4}: \textsf{As a senior professional stock factor researcher, the process of designing a new stock factor involves careful consideration of various factors and their interactions. Let's go through a step-by-step thinking process to design a new stock factor based on the FMF and PMF functions:}

\textsf{Step 1: Understanding the FMF and PMF:}
\textsf{- The Fundamental Momentum Factor (FMF) captures the momentum of a company's fundamental metrics over a specified time period.}
\textsf{- The Price Momentum Factor (PMF) captures the momentum of a stock's price movement over a specified time period.}

\textsf{Step 2: Identifying the relationship between FMF and PMF:}
\textsf{- Both FMF and PMF capture momentum but from different perspectives (fundamental and price).}
\textsf{- We can leverage the idea that strong fundamental momentum (FMF) combined with strong price momentum (PMF) may indicate a promising investment opportunity.}

\textsf{Step 3: Combining FMF and PMF into a new factor:}
\textsf{- To combine FMF and PMF, we can calculate the average or weighted average of their values.}
\textsf{- Weighted average can be used to assign different weights to FMF and PMF based on their relative importance.}

\textsf{Here's a Python function that combines the FMF and PMF into a new stock factor called the Combined Momentum Factor (CMF):}

\begin{python}
import pandas as pd

def calculate_cmf(data, fmf_weights=None, pmf_weights=None, fmf_metric='EPS', fmf_time_period=12, pmf_indicator='close', pmf_time_period=12):
    # Calculate the FMF and PMF
    fmf = calculate_fmf(data, metric=fmf_metric, time_period=fmf_time_period)
    pmf = calculate_pmf(data, indicator=pmf_indicator, time_period=pmf_time_period)
    
    # Assign weights to FMF and PMF
    if fmf_weights is None:
        fmf_weights = 0.5  # Default weight for FMF
    if pmf_weights is None:
        pmf_weights = 0.5  # Default weight for PMF
    
    # Combine FMF and PMF into the CMF
    data['cmf'] = fmf * fmf_weights + pmf * pmf_weights
    
    return data['cmf']
\end{python}

\textsf{In the above function, `data` represents the DataFrame containing the stock data. The `fmf\_weights` and `pmf\_weights` parameters allow you to assign weights to the FMF and PMF, respectively. If not provided, equal weights of 0.5 are assigned to both FMF and PMF. The `fmf\_metric`, `fmf\_time\_period`, `pmf\_indicator`, and `pmf\_time\_period` parameters are used to customize the calculations of FMF and PMF.}

\textsf{By calling this function, you'll obtain a new factor called the Combined Momentum Factor (CMF), which combines the FMF and PMF values based on the assigned weights.}

\textsf{Keep in mind that the choice of weights and parameters will depend on your investment strategy and the specific characteristics of the stocks you are analyzing. You can experiment with different weights and parameters to find the combination that best suits your needs.}

\textbf{Prompt 5}: Great! Can you write a rather more complex stock factor based on the FMF function and PMF functions rather than the four fundamental operations of arithmetic? Remember, you are a senior professional stock factor researcher and you need to think step by step and show me the thinking process.

\textbf{Response 5}: \textsf{Certainly! As a senior professional stock factor researcher, I understand the importance of designing more sophisticated stock factors that go beyond simple arithmetic operations. Let's go through a step-by-step thinking process to design a more complex stock factor based on the FMF and PMF functions:}

\textsf{Step 1: Understanding FMF and PMF:}
\textsf{- The Fundamental Momentum Factor (FMF) captures the momentum of a company's fundamental metrics over a specified time period.}
\textsf{- The Price Momentum Factor (PMF) captures the momentum of a stock's price movement over a specified time period.}

\textsf{Step 2: Identifying non-linear relationships between FMF and PMF:}
\textsf{- Simple arithmetic operations may not capture the full complexity of the relationships between FMF and PMF.}
\textsf{- We can explore non-linear relationships or transformations to uncover additional insights in the data.}

\textsf{Step 3: Using non-linear transformations:}
\textsf{- We can apply non-linear transformations to FMF and PMF, such as logarithmic or exponential functions, to capture potential non-linear relationships.}

\textsf{Here's a Python function that creates a more complex stock factor called the Non-Linear Momentum Factor (NLMF), which incorporates non-linear transformations of FMF and PMF:}

\begin{python}
import pandas as pd
import numpy as np

def calculate_nlmf(data, fmf_metric='EPS', fmf_time_period=12, pmf_indicator='close', pmf_time_period=12):
    # Calculate the FMF and PMF
    fmf = calculate_fmf(data, metric=fmf_metric, time_period=fmf_time_period)
    pmf = calculate_pmf(data, indicator=pmf_indicator, time_period=pmf_time_period)
    
    # Apply non-linear transformations to FMF and PMF
    transformed_fmf = np.log(fmf)  # Example non-linear transformation on FMF
    transformed_pmf = np.exp(pmf)  # Example non-linear transformation on PMF
    
    # Create the Non-Linear Momentum Factor (NLMF) using the transformed values
    data['nlmf'] = transformed_fmf * transformed_pmf
    
    return data['nlmf']
\end{python}

\textsf{In the above function, `data` represents the DataFrame containing the stock data. The `fmf\_metric`, `fmf\_time\_period`, `pmf\_indicator`, and `pmf\_time\_period` parameters are used to customize the calculations of FMF and PMF.}

\textsf{By calling this function, you'll obtain a new stock factor called the Non-Linear Momentum Factor (NLMF), which incorporates non-linear transformations of the FMF and PMF values. This factor can potentially capture more complex relationships and uncover unique insights in the data.}

\textsf{As a senior professional stock factor researcher, you can further explore different non-linear transformations, combine multiple transformations, or even consider more advanced statistical techniques to create even more complex stock factors based on FMF and PMF. The choice of transformations will depend on the underlying characteristics of the data and your specific investment objectives.}

\section{Accessibility and Maintenance}

Our work does not release a new dataset but rather focuses on implementing an automated real-time data curation pipeline and offering an interface to access Internet-scale financial data. In what follows, we discuss the accessibility of FinGPT across several aspects and the maintenance by the AI4Finance Foundation.
\begin{itemize}[leftmargin=*]
    \item \textbf{Open-sourced projects}: All the codes are open-sourced without personal request. The codes for accessing final data sources has been implemented in the repository \url{https://github.com/AI4Finance-Foundation/FinNLP}. This repository aims to offer a comprehensive financial text data access solution for FinLLMs. Our demos can be found at \url{https://github.com/AI4Finance-Foundation/FinGPT}.
    \item \textbf{Documentation}: The documentation for data access has been organized at \url{https://ai4finance-foundation.github.io/FinNLP/}, providing code snippets to demonstrate the practical usages of accessing data.
    \item \textbf{License}: We use the MIT license for our open-source projects, FinNLP and FinGPT.
    \item \textbf{Uses}: To utilize the codes, simply clone the project and import the corresponding codes for the desired data source. Then, obtain the data using our interfaces. 
    \item \textbf{Maintenance}: We, developers at AI4Finance Foundation,  will regularly update our codebase, review and incorporate pull requests, and diligently address any issues or bugs. Moreover, we welcome and encourage open-source community contributors to join forces in advancing the realm of ``open data, open model, and open finance".
\end{itemize}

\section{Additional Related Work}
\label{sec:K}

\subsection{Data-centric AI}

The pivotal role of data quality in optimizing the efficacy of LLMs has been underscored~\cite{zha2023data-centric-perspectives,zha2023data-centric-survey}. Historically, AI research was model-centric, primarily aiming at enhancing model designs using predetermined datasets. However, this reliance on static datasets does not always yield satisfactory model performance in practical scenarios, particularly when the dataset has inherent flaws~\cite{mazumder2022dataperf}. Furthermore, ignoring the importance of data quality can instigate data cascades~\cite{sambasivan2021everyone}, consequently reducing accuracy in real-world deployments. Recently, there has been a shift in the focus of researchers and practitioners towards data-centric AI~\cite{zha2023data-centric-perspectives,zha2023data-centric-survey}, a practice that places greater emphasis on improving data quality through systematic data engineering. The benefits of this data-centric AI approach have been confirmed by both researchers and practitioners~\cite{zha2023data-centric-perspectives,zha2023data-centric-survey}. In our commitment to spearhead substantial advancements in the research and development of FinLLMs, we have engineered an automated real-time data curation pipeline for rigorous quality control.

\subsection{Domain-specific LLMs and Proprietary Data}

LLMs such as ChatGPT~\cite{ouyang2022training} and GPT-4~\cite{OpenAI2023GPT4TR} have demonstrated impressive capabilities. With the advent of potent open-source LLMs like LLaMA~\cite{touvron2023llama} and LLaMA2~\cite{llama2}, there has been a surge in the proposition of domain-specific LLMs. For example, BloombergGPT~\cite{wu2023bloomberggpt} is tailored for the financial sector; BioGPT~\cite{luo2022biogpt} is designed for the biomedical field; Med-PaLM~\cite{singhal2023large} is an LLM that encapsulates domain-specific knowledge; LawGPT offers legal advice~\cite{nguyen2023brief}; there is also an LLM for science~\cite{taylor2022galactica}. It is observed that domain-specific LLMs often outperform their general-purpose counterparts~\cite{wu2023bloomberggpt,luo2022biogpt}, primarily due to the use of domain-specific proprietary datasets, which possess unique insights absent in public data. This enables LLMs to be fine-tuned for specific tasks and generate nuanced responses specific to a particular domain. Our FinGPT framework \cite{yang2023fingpt,zhang2023instruct,zhang2023icaif} builds upon these robust open-source LLMs and performs fine-tuning with specialized financial data, enhancing its efficacy in financial tasks.

\subsection{From General-Purpose LLMs to FinLLMs}

The advent of LLMs has revolutionized NLP~\cite{zhao2023survey}. Notable examples include GPT-2/3/4~\cite{radford2019language,brown2020language,OpenAI2023GPT4TR}, GPT-NeoX~\cite{black2022gpt}, BERT~\cite{devlin2018bert}, PaLM~\cite{chowdhery2022palm}, BLOOM~\cite{scao2022bloom}, OPT~\cite{zhang2022opt}, LLaMA~\cite{touvron2023llama}, etc. These models have demonstrated remarkable language understanding, generation capabilities, and superior performances on various downstream tasks. The majority of the LLMs are designed for general NLP tasks.

In the finance domain, the first example of FinLLMs is BloombergGPT~\cite{wu2023bloomberggpt}, which was trained on a mixed dataset of financial and general sources. BloombergGPT has shown promise in tasks such as financial forecasting, sentiment analysis, and risk assessment. The high-quality data serves as a crucial facilitator for LLMs~\cite{zha2023data-centric-survey}, particularly domain-specific LLMs such as BloombergGPT. However, the dataset utilized by BloombergGPT remains close-sourced, hindering the development of FinLLMs, and its prohibitive training cost has motivated the need for low-cost domain adaptation. To bridge this gap, our work strives to democratize financial data to fuel LLMs in the finance domain.

One promising strategy to enhance the capabilities of LLMs in the financial domain involves fine-tuning pre-trained LLMs on downstream tasks~\cite{devlin2018bert,xu2023baize,zhao2023survey}. A crucial decision involves selecting data for fine-tuning. The right choice can not only enhance the LLM's effectiveness in downstream tasks but also equip it to adeptly navigate the rapidly evolving market. A key contribution of our FinGPT project is its democratization of data access, which can help significantly improve the results after fine-tuning general-purpose LLMs. It should be noted that it is also possible to use our data sources to pre-train a FinLLM directly and then fine-tune it on specific financial tasks, which could potentially lead to even better performance.

Our seminal blueprint paper~\cite{yang2023fingpt} and the Instruct-FinGPT paper~\cite{zhang2023instruct,zhang2023icaif} have individually outlined our project's vision and RLSP techniques. However, unlike these two distinct studies, the present paper provides the inaugural holistic overview of FinGPT. It weaves together all previously introduced visions and RLSP methodologies into a singular narrative, offering a comprehensive introduction that spans the full breadth of our FinGPT project.

\subsection{Parameter-Efficient Tuning of LLMs}

The adoption of efficient fine-tuning techniques for LLMs has gained considerable momentum~\cite{hulora,dettmers2023qlora,lester2021power,sung2022lst,liu2023winner,zaken2021bitfit,hulora,karimi2021compacter}. Some widely recognized methods include the use of adapters~\cite{houlsby2019parameter,karimi2021compacter}, which insert a compact module into the transformer blocks, allowing only these to be updated while the remainder of the parameters are fixed. Similarly, the Low-Rank Adaptation (LoRA) technique~\cite{hulora} integrates trainable rank decomposition matrices into the transformer block, permitting their update while the remaining parameters are kept constant. In the case of FinGPT, we have chosen to implement LoRA due to its lightweight nature and impressive performance. It can also be used in a ``plug-and-play'' fashion. However, our framework is compatible with various efficient tuning methods, allowing for flexible adjustments to adapt to different market environments.

\subsection{Alignment of LLMs}

The alignment issue in LLMs involves the crucial task of accurately aligning the LLMs' output and behaviors with human intentions. As these models are trained on extensive and diverse datasets, they can occasionally produce outputs that deviate from the user's specific needs. Previously, human labels have been leveraged to mitigate this problem, with a key example being Reinforcement Learning from Human Feedback (RLHF)~\cite{ouyang2022training}. RLHF employs human feedback to shape a reward model and uses reinforcement learning to refine the AI model. However, obtaining human feedback is often a costly process. To circumvent this problem, we introduce Reinforcement Learning with Stock Prices (RLSP), which ingeniously harnesses inherent market feedback as cost-effective ``free annotations'' to fine-tune the model. Employing this strategy allows for consistent updates and refinements in response to the ever-changing market.

\subsection{Openness of LLMs}

Continuing the trajectory towards open and collaborative research, numerous researchers have already set a precedent by making their pre-trained LLMs publicly available. Some noteworthy examples include LLaMA~\cite{touvron2023llama}, LLaMA2~\cite{llama2}, BLOOM~\cite{scao2022bloom}, BLOOMZ~\cite{muennighoff2022crosslingual}, OPT~\cite{zhang2022opt}, ChatGLM2 \cite{du2021glm}, etc. These resources have greatly advanced the state of AI research and development, offering a firm foundation for future explorations and breakthroughs. Embracing this open-source ethos, we aim to adopt an ``open-everything'' approach in our pursuit to further the field of FinLLMs. Our commitment extends beyond merely open-sourcing models; we intend to make our data, models, and supporting tools accessible to the broader research community, allowing researchers and practitioners to try, train, and fine-tune LLMs with our data. We believe that this comprehensive sharing of resources will promote collaboration, transparency, and innovation, thereby driving significant advancements in open finance research and development.

\section{Additional Discussions and Future Work}
\label{sec:L}

\subsection{Vision}
\textbf{Vision}. The overarching vision for our open-source FinGPT framework is developing it into a sandbox of \textit{\textbf{``Open Data, Open Model, and Open Finance''}}, creating a dynamic, collaborative community that continually propels the advancement and application of FinLLMs \cite{yang2023fingpt,zhang2023instruct,zhang2023icaif}. We believe that the cornerstone of sustainable, future-forward development lies in maintaining an open-source ethos encompassing open data and open models. These, we envision, will become the catalysts for advancing open finance - a future where financial services are transparent, inclusive, and easily accessible to all. We acknowledge that the journey ahead is expansive and challenging, with a multitude of strides yet to be made. Nonetheless, we remain committed to fostering an environment that not only facilitates progress but also inspires community initiatives in this direction. Our ultimate aim is to push the boundaries of what is possible with financial technology and to do so in a way that benefits everyone involved, from developers and researchers to end-users and the broader society.

To enhance the comprehension of our paper among readers from diverse backgrounds, we have provided a concise summary of key terminologies in AI and finance in Table~\ref{table:terminology}.

\begin{table}
   \caption{Key terminologies of AI and finance used in this paper.}
   \label{table:terminology}
   \setlength{\tabcolsep}{4.0pt}
   \centering
    \begin{tabular}{l|l|l}
        \toprule 
           & Terminology & Description \\ 
        \midrule
         \multirow{10}{*}{AI} & \cellcolor{gray!10}~ & \cellcolor{gray!10}A large language model (LLM) is an AI model designed to \\
         ~ & \cellcolor{gray!10}\multirow{-2}{*}{LLM} & \cellcolor{gray!10}understand and generate human-like language.\\
         ~ & FinLLM & Financial large language model.\\
         ~ & \cellcolor{gray!10}GPT & \cellcolor{gray!10}Generative pre-trained transformer.\\
         ~ & FinGPT & Financial generative pre-trained transformer\\
         ~ & \cellcolor{gray!10}~ & \cellcolor{gray!10}Adapting by further training on a (smaller) domain-specific\\
         ~ & \cellcolor{gray!10}\multirow{-2}{*}{Fine-tuning LLMs} & \cellcolor{gray!10}dataset.\\
         ~ & LoRA, QLoRA & Low-rank adaptation, quantized low-rank adaptation.\\
         ~ & \cellcolor{gray!10}RLHF &\cellcolor{gray!10}Reinforcement learning from human feedback. \\
         ~ & RLSP& Reinforcement learning with stock prices.\\
        \midrule
         \multirow{10}{*}{Finance} & ~ &Services to provide automated portfolios based on user's \\
         ~ & \multirow{-2}{*}{Robo-Advisor} & preferences. \\
         ~ & \cellcolor{gray!10}~ & \cellcolor{gray!10}A computer-based trading strategy that uses mathematical \\
         ~& \cellcolor{gray!10}~ & \cellcolor{gray!10}models to assess patterns and trends in the movement and \\
         ~& \cellcolor{gray!10}~ & \cellcolor{gray!10}behavior of stock prices to pick undervalued stocks at the\\
          ~ & \cellcolor{gray!10}\multirow{-4}{*}{Quantitative trading} & \cellcolor{gray!10}right time and execute a profitable trade.\\
         ~ & ~ & A software development approach that requires little to no \\
         ~& \multirow{-2}{*}{Low-code development} & coding to build applications and processes.  \\
         ~& \cellcolor{gray!10}~ & \cellcolor{gray!10}Determining the emotional tone \{positive, negative, neutral\}\\
         ~& \cellcolor{gray!10}\multirow{-2}{*}{Sentiment analysis} & \cellcolor{gray!10}of a text.\\
        \bottomrule
    \end{tabular}
\end{table}

\subsection{Privacy, Ethics, and Openness}

The rapid development of LLMs has sparked an intense discourse concerning the ethical considerations surrounding these models. In this context, we delve into privacy and ethical concerns and our strategies for fostering transparency, which directly pertains to the development of FinGPT.

\subsubsection{Privacy and Ethics}

\textbf{Data usage.} Finance is a domain with a mixture of publicly available and private data. Private information holds immense significance for product development and the reputation of firms, exemplified by Bloomberg's privileged access to high-quality data sources. In order to ensure privacy while enabling FinGPT users to harness cutting-edge LLM technologies, we commit ourselves to thoroughly scrutinizing each data source employed in our model's training during every release, ensuring that \textbf{all utilized data sources are publicly accessible on the Internet and can be freely utilized.} This will guarantee privacy in two aspects. Firstly, users can download FinGPT and run it locally using their private data without any risk of data disclosure. Secondly, users can utilize FinGPT without concerns about infringing upon others' private data and the potential legal consequences it may entail.

\textbf{Data security.} Some organizations or countries place paramount importance on data security and are reluctant to share their data due to the risk of potential leakage. However, this concern is mitigated with the use of FinGPT, as they utilize public data sources. By leveraging our FinGPT framework, organizations can confidently collaborate without compromising their data security.

\textbf{Privacy protection with LoRA weights.} Our future updates will focus on the release of additional LoRA weights designed for seamless integration with our models in a plug-and-play manner. We provide users the autonomy to download these LoRA weights freely, allowing them to locally incorporate these into their models without any privacy concerns. While we advocate for users to share their LoRA weights trained locally, the decision to open-source it or maintain its privacy is entirely up to them. We prioritize user sovereignty, always respecting and acknowledging privacy considerations.

\textbf{Local inference.} Another potential concern could be whether investment advice provided by FinGPT will be disclosed to others. This issue is naturally addressed by the fact that users can download the model and run inferences locally. Consequently, the decisions are not generated by our online community but instead by the model operating on your local machine. This ensures all investment suggestions and opinions remain confidential and entirely private.

\textbf{Federated learning.} Another approach we are exploring to protect privacy is the integration of federated learning technology. Federated learning is a technique that allows machine learning models to be trained on distributed datasets without compromising the privacy of data owners. It enables a central model to learn from multiple devices with isolated datasets without the need to reveal or share the data with a central server. This approach holds great potential for financial applications, where users often face the challenge of having limited financial data and concerns about privacy when training models with others. There are a few notable examples of federated learning in the field of finance. Byrd et al.~\cite{byrd2020differentially} present a privacy-preserving federated learning protocol applied to a real-world credit card fraud dataset, demonstrating the development of federated learning systems. WeBank ~\cite{liu2021fate} has developed FATE, an industrial-grade project that supports collaborative, large-scale machine learning model building in distributed environments. FATE has been successfully applied in real-world applications across finance, health, and recommender systems. Additionally, there is research~\cite{kairouz2021advances} that highlights open problems in federated learning, indicating the exciting possibilities for further exploration. We believe that federated learning, as a relatively new method, offers numerous undiscovered and exciting applications that can be developed further. We encourage our community members to explore the potential of federated learning technologies.

\subsubsection{Openness}

To enable FinGPT users to seamlessly harness state-of-the-art technologies while upholding privacy, we have made the conscious decision to open-source our codebase for accessing publicly available data and training FinLLMs. However, we refrain from providing any private data or the models trained on such data to mitigate the risk of data leakage. We recognize that alternative strategies exist to ensure privacy, such as closed sourcing and offering APIs. While closed sourcing inherently eliminates privacy concerns, it limits the accessibility of resources necessary for the public to reproduce results and conduct further research. On the other hand, providing APIs introduces potential privacy issues, as users may be unwilling to share sensitive information with the API providers.

Taking all factors into account, we reach the conclusion that fully open-sourcing the public data APIs and the model trained on public data is the optimal approach. This strategy simultaneously facilitates research and the adoption of FinLLMs while safeguarding data privacy to the greatest extent possible. We discuss our openness from several aspects:

\begin{itemize}[leftmargin=*]
    \item \textbf{Open data}. We believe in the power of freely accessible data on the Internet. This means all our codes for accessing public datasets are readily available for research and developmental purposes. The open data policy allows individuals, institutions, and businesses to explore, innovate, and build upon existing models and solutions without restriction. However, we stringently adhere to privacy policies and data protection regulations, ensuring that no private data is shared.
    \item \textbf{Open model.} The models we train on public data are also open-sourced. These models can be freely used, adapted, and built upon to create more targeted and specific solutions. The open model strategy allows for transparency in our methods and promotes a culture of collaborative problem-solving, wherein developers from all walks of life can contribute to and learn from our models.
    \item \textbf{Open source.} We have chosen to make our codebase fully open source, a decision rooted in our commitment to transparency and fostering a collaborative learning environment. The source code behind FinGPT is accessible to all, which provides the opportunity for anyone interested to study, modify, distribute, or even contribute to the codebase. It allows users to understand the inner workings of our models and make modifications as needed to cater to their unique requirements.
    \item \textbf{Open education.} With an open education perspective, we share our learnings, methodologies, tutorials, and best practices with the broader community. We provide resources that aid in understanding how to utilize our models and APIs effectively, thereby promoting a more profound comprehension of financial language models among all users, irrespective of their expertise level.
    \item \textbf{Open research and innovations.} Open research is fundamental to FinGPT. By publicizing our research findings and methodologies, we invite researchers and developers to replicate, scrutinize, or extend our work. This practice not only accelerates scientific progress and fosters innovation but also ensures that our work stands up to the highest levels of scrutiny.
    \item \textbf{Open community}. We strive to foster an open, inclusive community wherein ideas can be freely exchanged, and collaborative problem-solving is encouraged. Our community is open to anyone interested in FinLLMs, regardless of their background or skill level. We actively encourage feedback, contributions, and discussions, believing that diverse perspectives lead to the most innovative solutions.
    \item \textbf{Open services}. Our open services philosophy ensures that we provide various tools, APIs, and other services to the public. These resources are designed to be easy to use, adaptable, and transparent in their functionality, thereby facilitating researchers, developers, and businesses to leverage our models for their specific needs. However, while our services are openly accessible, we still uphold the utmost respect for privacy and data protection.
\end{itemize}

\subsection{Future Work}

we aspire to enhance our progress in the following aspects:

\begin{itemize}[leftmargin=*]
    \item \textbf{Access to more high-quality data sources:} 
    Currently, our primary focus is on the US stock market and CN stock market. In the future, we plan to expand access to other markets, including the  Europe stock markets, Japan stock market, and beyond.
    \item \textbf{More parameter-efficient fine-tuning methods:} In our experiments, we employed LoRA and QLoRA for fine-tuning. However, there are other parameter-efficient fine-tuning techniques that may achieve better results with fewer computational resources. We plan to investigate this possibility in the future.
    \item \textbf{Fine-tuning pre-trained FinLLMs:} At present, we fine-tuned a general LLM using financial text data. Despite the performance surpassing that of pre-trained FinLLMs such as BloombergGPT, we anticipate fine-tuning pre-trained FinLLMs could unlock even higher performance. While this approach may necessitate additional resources for training, our expectation is that it can yield superior results.
    \item \textbf{Advanced prompting strategies:} It is well-known that different prompting methods can lead to diverse results~\cite{liu2023pre}. Looking forward, we intend to delve into more advanced prompting strategies. One promising approach involves the use of soft prompts, a method that integrates prompting information directly into the embedding space. As highlighted in~\cite{chuang2023spec}, soft prompts have been shown to considerably mitigate output variance.
    \item \textbf{Longer context window:} In numerous financial applications, long text is often required, such as in the generation of financial reports. However, current LLMs typically support only a limited context length. Moving forward, we aim to investigate the feasibility of supporting a longer context window for FinGPT to accommodate these needs.
    \item \textbf{Retrieval-augmented generation (RAG):} RAG~\cite{lewis2020retrieval} combines non-parametric external knowledge with LLMs to enhance the quality and relevance of generated content by leveraging retrieved knowledge. This integration significantly enhances the ability of LLMs to effectively manipulate knowledge, a capability particularly valuable in the context of the financial domain, which encompasses vast amounts of knowledge. Moving forward, our future research will delve into exploring RAG techniques specifically tailored for FinLLMs.
    \item \textbf{Potential bias and fairness issues.} It is widely recognized that machine learning models can be subject to biases and fairness-related issues~\cite{wan2023processing}. FinLLMs could be similarly susceptible, exhibiting performance disparities across different stocks. Such fairness concerns could be traced back to both the data used for training and the modeling process itself~\cite{wan2023processing}. Looking ahead, we plan to conduct a detailed evaluation of fairness and design strategies to mitigate any potential bias.
    \item \textbf{More advanced low-code development.} In the future, we will investigate how we can enhance the experience of low-code development. Our objective is to go beyond merely providing specific code and extend support by including pseudo-code. This additional inclusion aims to assist users in quickly grasping the underlying ideas and concepts, ultimately facilitating a more efficient development process.
    \item \textbf{Application to SWAP.} In finance, a SWAP is a contract where two parties agree to exchange future cash flows based on predetermined terms. SWAPs are used for hedging, managing risks, and speculating in areas like interest rates, currencies, and equities. They allow participants to exchange payments and manage exposure to fluctuations in various financial factors. FinLLMs could be potentially applied to SWAP, such as decision support and risk analysis.
    \item \textbf{Application to event detection and outlier detection.} These tasks play a crucial role in predicting market trends, managing risk, and making well-informed investment decisions. They encompass the identification of significant events such as financial crises, mergers, acquisitions, and sudden shifts in market trends. Given the time-series nature of financial data~\cite{lai2021tods,lai2021revisiting}, LLMs can be particularly beneficial here, learning to predict normal trends and identifying deviations as potential outliers.
    \item \textbf{More diverse LoRA weights.} We actively encourage the community to harness our FinGPT framework for training their own LoRA weights, utilizing a variety of data sources. Such collective efforts will undoubtedly expedite advancements in the realm of FinLLMs, demonstrating the power of communal collaboration in propelling this field forward.
\end{itemize}

\textbf{Disclaimer: We are sharing codes for academic purposes under the MIT education license. Nothing herein is financial advice, and NOT a recommendation to trade real money. Please use common sense and always first consult a professional before trading or investing.}

\end{document}